\documentclass{article} 
\usepackage{colm2024_conference}

\usepackage{booktabs}
\usepackage{graphicx}
\usepackage{enumitem}
\usepackage{wrapfig}
\usepackage{algorithm}
\usepackage{algpseudocode}
\usepackage{natbib}
\usepackage{makecell}
\usepackage{booktabs}
\usepackage{bbm}
\usepackage{array}
\usepackage{amsmath} 
\usepackage{amssymb}
\usepackage{amsfonts}
\usepackage{multirow}
\usepackage{verbatim}
\usepackage{caption}
\usepackage{longtable}
\usepackage{supertabular}
\usepackage{hyperref}
\usepackage{CJKutf8}
\usepackage[utf8]{inputenc} 
\usepackage[T1]{fontenc} 
\usepackage[french,vietnamese,mongolian,greek,english]{babel}
\usepackage{pifont}
\usepackage{afterpage}
\usepackage{enumitem}
\usepackage{tablefootnote}
\usepackage{xspace}
\usepackage{textcomp}
\usepackage{makecell}
\usepackage{lscape} 
\usepackage{siunitx}
\usepackage{listings}
\usepackage{xcolor}
\usepackage{adjustbox}
\usepackage[table]{xcolor}
\definecolor{dt}{gray}{0.7}

\newcommand{\cmark}{\ding{51}}%
\newcommand{\xmark}{\ding{55}}%

\newcommand{\officialscore}{\textsuperscript{\scriptsize\ensuremath{\dagger}}}
\newcommand{\paperascore}{\textsuperscript{\scriptsize\ensuremath{\ddagger}}}
\newcommand{\paperbscore}{\textsuperscript{\scriptsize\ensuremath{\S}}}
\newcommand{\papercscore}{\textsuperscript{\scriptsize\ensuremath{\P}}}

\usepackage{pifont}       
\usepackage{bbding}       
\usepackage{fontawesome}

\usepackage{scrextend}

\usepackage{tgpagella}
\usepackage{latexsym}
\usepackage[T1]{fontenc}
\usepackage[utf8]{inputenc}
\usepackage{microtype}
\definecolor{mydarkblue}{rgb}{0,0.08,0.45}
\definecolor{citecolor}{HTML}{0071BC}
\usepackage{url}            
\usepackage{nicefrac}       
\usepackage{changepage}
\usepackage{xargs}          
\usepackage{wrapfig,lipsum,booktabs}
\usepackage{longtable}
\usepackage{subcaption}
\usepackage{endnotes}

\usepackage{titlesec}

\renewcommand{\theparagraph}{\thesubsubsection.\arabic{paragraph}}

\titleformat{\paragraph}
  {\normalfont\normalsize\bfseries}
  {\theparagraph}
  {1em}
  {}

\titlespacing*{\paragraph}
  {0pt}
  {1.25ex plus 1ex minus .2ex}
  {0.8ex}

\usepackage[most]{tcolorbox}
\usepackage{xcolor}
\usepackage{listings}

\lstdefinelanguage{json}{
  morestring=[b]",
  stringstyle=\color{blue!45!black},
  showstringspaces=false
}

\usepackage{pgfplots}
\usetikzlibrary{pgfplots.groupplots}
\pgfplotsset{compat=1.3}
\usepackage{tikz}
\usetikzlibrary{patterns}

\usepackage[most]{tcolorbox}
\usepackage{fvextra}
\usepackage{graphicx}
\usepackage[capitalize,noabbrev]{cleveref}
\crefname{section}{Section}{\S\S}
\Crefname{section}{Section}{\S\S}
\crefname{table}{Table}{Tables}
\crefname{figure}{Figure}{Figures}
\crefname{algorithm}{Algorithm}{}
\crefname{equation}{eq.}{}
\crefname{appendix}{Appendix}{}
\crefformat{section}{Section #2#1#3}
\usepackage{multicol}
\usepackage{fancyvrb} 
\usepackage{tcolorbox}
\newsavebox{\myverbcontent}
\usepackage{titlesec}
\titleformat*{\section}{\large\bfseries}

\usepackage{nicematrix} 
\usepackage{arydshln}

\makeatletter
\DeclareRobustCommand\onedot{\futurelet\@let@token\@onedot}
\def\@onedot{\ifx\@let@token.\else.\null\fi\xspace}

\title{S1-DeepResearch: Beyond Search, \\Toward Real-World Long-Horizon Research Agents}

\author{
\bf XScience Lab\\
Wenge AI}

\begin{document}

\maketitle

\begin{abstract} 
Deep research agents aim to solve complex knowledge-intensive tasks through long-horizon planning, evidence gathering, reasoning, and report generation. While recent progress in search agents has demonstrated strong capabilities in information retrieval and answer verification, most existing training datasets remain search-centric, focusing primarily on closed-ended question answering and information localization. As a result, they mainly train information-seeking behavior while providing limited coverage of key deep research capabilities, including evidence integration, knowledge synthesis, planning and decision making, file understanding, and structured report generation. In this work, we propose a unified trajectory construction paradigm for deep research agents that combines closed-ended QA and open-ended exploration. The proposed framework consists of graph-grounded task formulation, agentic trajectory rollout, and multi-dimensional trajectory verification, enabling scalable synthesis of high-quality agentic trajectories spanning long-chain complex reasoning, deep research instruction following, deep research report writing, file understanding and generation, and skills usage. Compared with existing search-oriented datasets, our synthesized trajectories place greater emphasis on knowledge synthesis, complex reasoning, and planning and decision making. S1-DeepResearch-32B achieves state-of-the-art performance among open-source models of comparable scale across 20 benchmarks spanning five capability dimensions, including complex reasoning, deep research instruction following, report generation, file understanding, and skills usage. On several challenging deep research benchmarks, it approaches the performance of leading proprietary frontier models. These results highlight the importance of jointly modeling information acquisition, knowledge synthesis, and planning-oriented agent behaviors for building effective deep research agents. To facilitate future research, we release both S1-DeepResearch-32B and S1-DeepResearch-15K, a collection of 15K high-quality agentic trajectories constructed using our framework.
\end{abstract}
\begin{figure*}[ht]
    \centering
    \includegraphics[width= 0.9\linewidth,height=0.3\textheight,trim=0 15 0 25,]{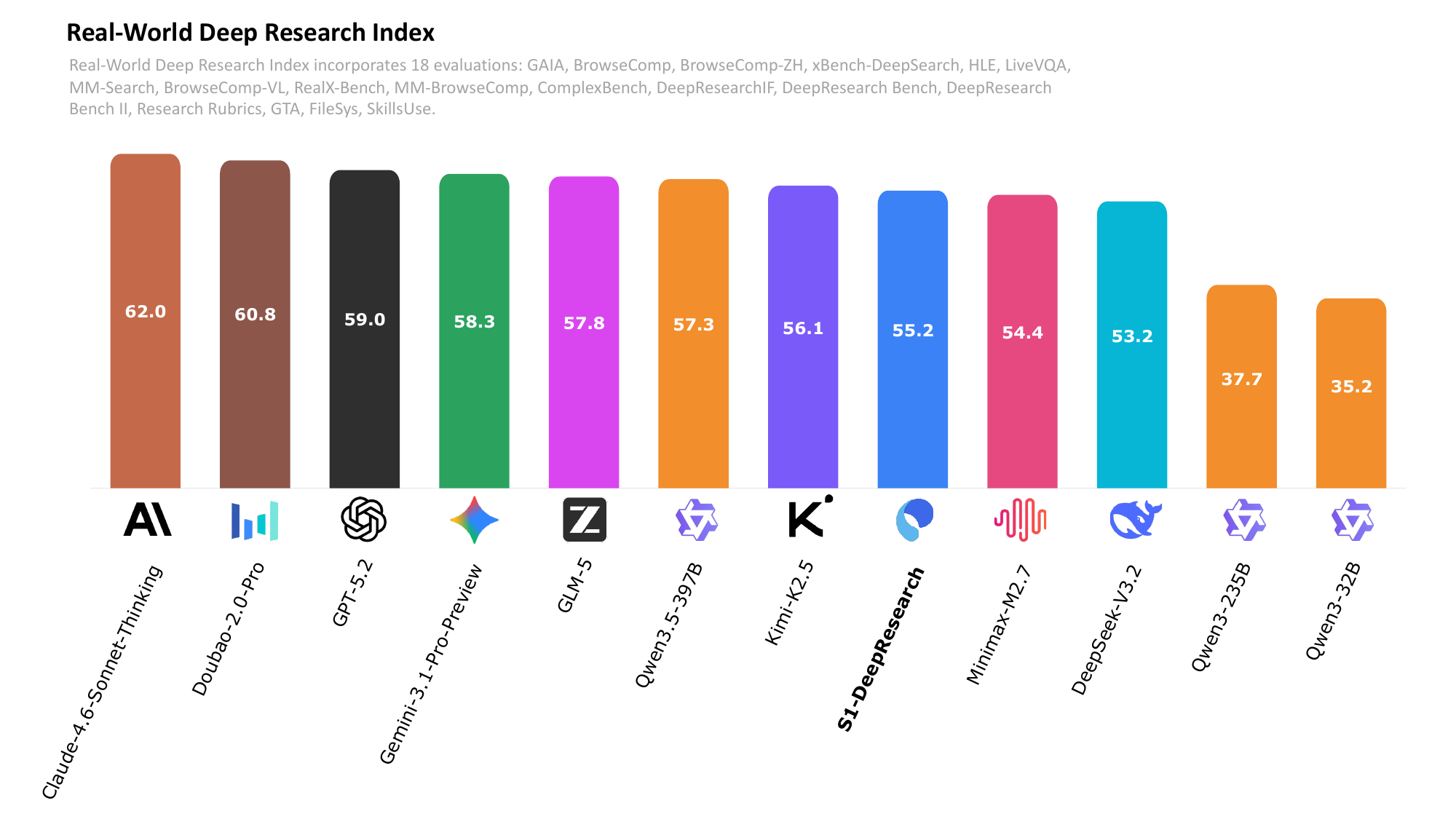}
    \caption{Average scores across five deep research capability dimensions.}
    \label{fig:overall_performance}
\end{figure*}

\newpage
\section{Introduction}

Large language models (LLMs) are expanding from static text generation toward agentic problem solving in real environments: instead of answering a single question, a model must plan over long interactions, call tools, gather evidence, and revise its behavior based on feedback \citep{nakano2021webgpt,yao2023react,schick2023toolformer}. This transition is especially important for deep research, where scientific research, industry analysis, and knowledge-intensive workflows often involve multi-stage goals, heterogeneous sources, and complex constraints. Such tasks require long-chain complex reasoning that keeps search, evidence aggregation, state maintenance, and result generation aligned. Deep research is therefore not equivalent to deep search: the latter focuses on locating and verifying information for determinate answers, while the former also requires building analysis frameworks for open-ended goals, resolving evidence conflicts, and producing defensible, citable, and deliverable research outputs.

Recent work on long-horizon search agents and open-ended research agents has shown that scalable task synthesis, tool-augmented interaction, and trajectory-based post-training can substantially improve models' information-seeking and agentic reasoning abilities \citep{chu2026redsearcher,du2026openseeker,gao2025beyond,liu2025webexplorer,li2026openresearcher,hu2025stepdeepresearch,yao2026oresearcher,huang2026visiondeepresearch,yao2026mmdeepresearch}. However, most existing training data remains search-centric, focusing primarily on closed-ended QA, information localization, and evidence retrieval. Such data is scalable and easy to verify, but it provides limited coverage of key deep research capabilities, including evidence integration, knowledge synthesis, planning and decision making, file understanding, and structured report generation.

We argue that the central bottleneck is the scarcity of high-quality agentic trajectories that are both scalable and faithful to real deep research. Closed-ended QA offers clear correctness signals and supports large-scale synthesis and filtering, but it captures only part of the research process. Open-ended exploration is closer to real research needs, where goals may be underspecified, evidence may be incomplete or conflicting, and multiple valid outputs may exist; yet such tasks are difficult to synthesize, automatically verify, and control. A useful data construction paradigm for deep research agents must therefore combine the verifiability of closed-ended tasks with the realism of open-ended exploration.

In this paper, we introduce S1-DeepResearch, an agentic model and data framework for deep research. We adopt a unified trajectory construction paradigm that combines closed-ended QA and open-ended exploration, consisting of graph-grounded task formulation, agentic trajectory rollout, and multi-dimensional trajectory verification. The resulting trajectories cover five capability dimensions: Long-chain Complex Reasoning, Deep Research Instruction Following, Deep Research Report Writing, File Understanding and Generation, and Skills Usage. Compared with search-oriented datasets, our trajectories place greater emphasis on knowledge synthesis, complex reasoning, planning and decision making, and deliverable-oriented generation.

Our contributions are threefold. First, we release S1-DeepResearch-32B and S1-DeepResearch-15K\footnote{Dataset: \url{https://huggingface.co/datasets/ScienceOne-AI/S1-DeepResearch-15k}}, a collection of 15K high-quality agentic trajectories constructed using our framework. Second, we propose a scalable trajectory construction paradigm that jointly models information acquisition, knowledge synthesis, and planning-oriented agent behaviors by combining closed-ended QA with open-ended exploration. Third, we conduct systematic evaluations across 20 benchmarks spanning five capability dimensions, where S1-DeepResearch-32B achieves state-of-the-art performance among open-source models of comparable scale and approaches leading proprietary frontier models on several challenging deep research benchmarks.

\section{Related Work}
\label{sec:related_work}

\subsection{System and Workflow-Driven Deep Research}

One line of deep research work completes complex research tasks through explicit system orchestration. Systems such as OpenAI Deep Research and Gemini Deep Research demonstrate the practical value of this route in realistic deep research scenarios, where models conduct multi-step planning, search, reading, analysis, and citation-backed long-form report generation for open-ended questions \citep{openai2025deepresearch,googledeepmind2026deepresearchmax}. Similarly, MindDR~\citep{minddr2026technical}, AI-Researcher~\citep{tang2025airesearcher}, and AI-Scientist~\citep{lu2024aiscientist} attempt to organize task decomposition, evidence retrieval, experiment execution, and paper writing into multi-stage or multi-agent workflows. The significance of these methods is that they show deep research is not merely information retrieval, but a complete process involving planning, evidence collection, information integration, tool execution, and result presentation.

Correspondingly, recent evaluations have also moved beyond final-answer accuracy toward the quality of complete research outputs. DeepResearch Bench, DeepResearch Bench II, and ResearchRubrics evaluate deep research agents through long-form reports, expert rubrics, citation quality, factual grounding, and report-level reasoning \citep{du2025deepresearchbench,li2026deepresearchbench2,sharma2025researchrubrics}; Vision-DeepResearch, VDR-Bench, and MM-DeepResearch further extend search-based reasoning to visual and textual evidence \citep{huang2026visiondeepresearch,zeng2026visiondeepresearchbenchmark,yao2026mmdeepresearch}. Together, these works characterize the task form of real deep research: models must not only find information, but also handle multi-source materials, cross-modal evidence, citation constraints, and open-ended report generation.

However, the capabilities of system and workflow-driven methods largely depend on external modules, toolchains, prompt/workflow orchestration, or multi-agent collaboration. They demonstrate what a complex research workflow should accomplish, but it is not always clear whether the underlying model has acquired transferable native research capability. Meanwhile, open-ended research evaluations better reflect realistic tasks, but they usually focus on final output quality rather than providing complete behavioral trajectories for model training. Therefore, building more native deep research agents requires further discussion of how to distill these complex research behaviors into high-quality data and internalize them into model capabilities.

\subsection{Agentic Models for Deep Research}

Another line of work aims to internalize planning, search, reasoning, tool use, and evidence integration into the model itself through agentic training. Tongyi-DeepResearch, Step-DeepResearch, O-Researcher, MiroThinker, REDSearcher, OpenSeeker, ASearcher, WebExplorer, and OpenResearcher improve long-horizon search and tool-augmented reasoning from different perspectives, including agentic mid-training, SFT, RL, verification mechanisms, and trajectory synthesis \citep{tongyi2025deepresearch,hu2025stepdeepresearch,yao2026oresearcher,miromind2026mirothinker,chu2026redsearcher,du2026openseeker,gao2025beyond,liu2025webexplorer,li2026openresearcher}. These model-centric agents show that high-quality trajectories and post-training can substantially improve native agentic capabilities, reducing the model's dependence on external workflows for complex information-seeking tasks.

Agentic trajectory data is the key foundation of this direction. Early work on tool use and web interaction has shown that models can learn to call external tools, browse webpages, and update subsequent actions based on observations through demonstrations or synthesized trajectories \citep{nakano2021webgpt,yao2023react,schick2023toolformer}. Recent long-horizon search trajectories further improve multi-turn search, evidence localization, and path planning. Their advantage lies in scalability and verifiability: tasks usually have relatively determinate target answers, and trajectory quality can be filtered by answer correctness or retrieved evidence. However, this also makes many existing trajectories closer to extractive search, where the model locates, extracts, and verifies existing facts from an external information space. Real deep research is closer to constructive exploration, where the model must organize analytical frameworks, form argumentative structures, and generate deliverable outputs under open-ended goals, incomplete evidence, conflicting sources, and evolving constraints.

Therefore, although existing model-centric agents have shown that long-horizon search and tool-use abilities can be improved through trajectory-based training, much of the training data is still centered on closed-ended QA or verifiable search tasks, making it better suited for information localization and answer verification. Meanwhile, complex instruction following, file understanding and generation, and skills usage required by deep research are often evaluated separately or within relatively short, closed, and task-specific workflows \citep{zhou2023instruction,jiang2024followbench,qin2024infobench,wen2024complexbenchmarking,qi2025agentif,li2026skillsbench,li2026agentskillos,han2026sweskillsbench}. In contrast, S1-DeepResearch-15K aims to cover both closed-ended QA and open-ended exploration through unified trajectory data, and further organizes five capability dimensions: Long-chain Complex Reasoning, Deep Research Instruction Following, Deep Research Report Writing, File Understanding and Generation, and Skills Usage. This provides the basis for the data construction method described in the following section.

\begin{table}[t]
\centering
\caption{
Comparison of explicit agentic capability coverage across deep research models. Capability coverage refers to capabilities explicitly supported, evaluated, or described in the corresponding technical reports or released systems.
Reported training recipes are included for context.
}
\label{tab:capability_comparison}
\setlength{\tabcolsep}{4.5pt}
\renewcommand{\arraystretch}{1.12}
\small
\begin{tabular}{lccccccccc}
\toprule
\multirow{2}{*}{\textbf{Model}} &
\multicolumn{6}{c}{\textbf{Capability Coverage}} &
\multicolumn{3}{c}{\textbf{Training Recipe}} \\
\cmidrule(lr){2-7}
\cmidrule(lr){8-10}
& \textbf{LHR-Text} & \textbf{LHR-MM}
& \textbf{Instr.} & \textbf{Report}
& \textbf{Doc.} & \textbf{Skill}
& \textbf{Mid} & \textbf{SFT} & \textbf{RL} \\
\midrule
Tongyi-DeepResearch      & \cmark & \xmark & \xmark & \xmark & \cmark & \xmark & \cmark & \cmark & \cmark \\
OpenSeeker-v1               & \cmark & \xmark & \xmark & \xmark & \xmark & \xmark & \xmark & \cmark & \xmark \\
OpenResearcher           & \cmark & \xmark & \xmark & \xmark & \xmark & \xmark & \xmark & \cmark & \xmark \\
MiroThinker-1.7      & \cmark & \xmark & \xmark & \cmark & \cmark & \xmark & \cmark & \cmark & \cmark \\
REDSearcher              & \cmark & \xmark & \xmark & \xmark & \cmark & \xmark & \cmark & \cmark & \cmark \\
REDSearcher-MM           & \cmark & \cmark & \xmark & \xmark & \xmark & \xmark & \cmark & \cmark & \cmark \\
Skywork-R1V4             & \cmark & \cmark & \xmark & \xmark & \cmark & \xmark & \xmark & \cmark & \xmark \\
Vision-DeepResearch      & \cmark & \cmark & \xmark & \xmark & \cmark & \xmark & \xmark & \cmark & \cmark \\
MM-DeepResearch          & \cmark & \cmark & \xmark & \xmark & \xmark & \xmark & \xmark & \cmark & \cmark \\
UniScientist             & \cmark & \xmark & \xmark & \cmark & \cmark & \xmark & \xmark & \cmark & \xmark \\
Step-DeepResearch        & \cmark & \xmark & \xmark & \cmark & \cmark & \xmark & \cmark & \cmark & \cmark \\
\textbf{S1-DeepResearch} & \cmark & \cmark & \cmark & \cmark & \cmark & \cmark & \xmark & \cmark & \xmark \\
\bottomrule
\end{tabular}

\vspace{2pt}
\footnotesize
\textit{Note:} \cmark and \xmark \,indicate whether a capability is explicitly covered. 
For training recipes, \cmark denotes a reported or used stage, while \xmark \,denotes not used or not publicly specified. 
LHR = long-horizon reasoning; MM = multimodal; Instr. = deep research instruction following; Doc. = document understanding and generation; Skill = dynamic skill orchestration.
\end{table}

\section{Agentic Data Construction System}

Constructing exploration trajectories with high complexity and strong verifiability is critical for enabling large language models (LLMs) to acquire Deep Research capabilities. To this end, we design an automated Agentic Data Construction System that simulates the reasoning, exploration, and iterative refinement process of human researchers when solving complex real-world problems. Through carefully designed execution environments and multi-stage filtering mechanisms, our system synthesizes high-quality training data characterized by advanced tool use, long-context reasoning, and logically consistent decision-making trajectories.

\subsection{Overview}

As discussed in Section~\ref{sec:related_work}, existing trajectory synthesis methods are mostly limited to Extractive Search, where agents primarily perform information retrieval and aggregation. Table~\ref{tab:capability_comparison} further summarizes this limitation from the perspective of explicit capability coverage. Existing deep research models are often specialized toward long-horizon search and reasoning, while providing limited systematic support for other capabilities required in realistic deep research scenarios, such as fine-grained instruction following, report generation, document understanding and generation, and dynamic skill orchestration. Reported training recipes are also included for context, since several specialized systems adopt additional mid-training or reinforcement learning stages.

\begin{figure*}[t]
    \centering
    \includegraphics[width= \linewidth,height=0.35\textheight,trim=0 0 0 0,]{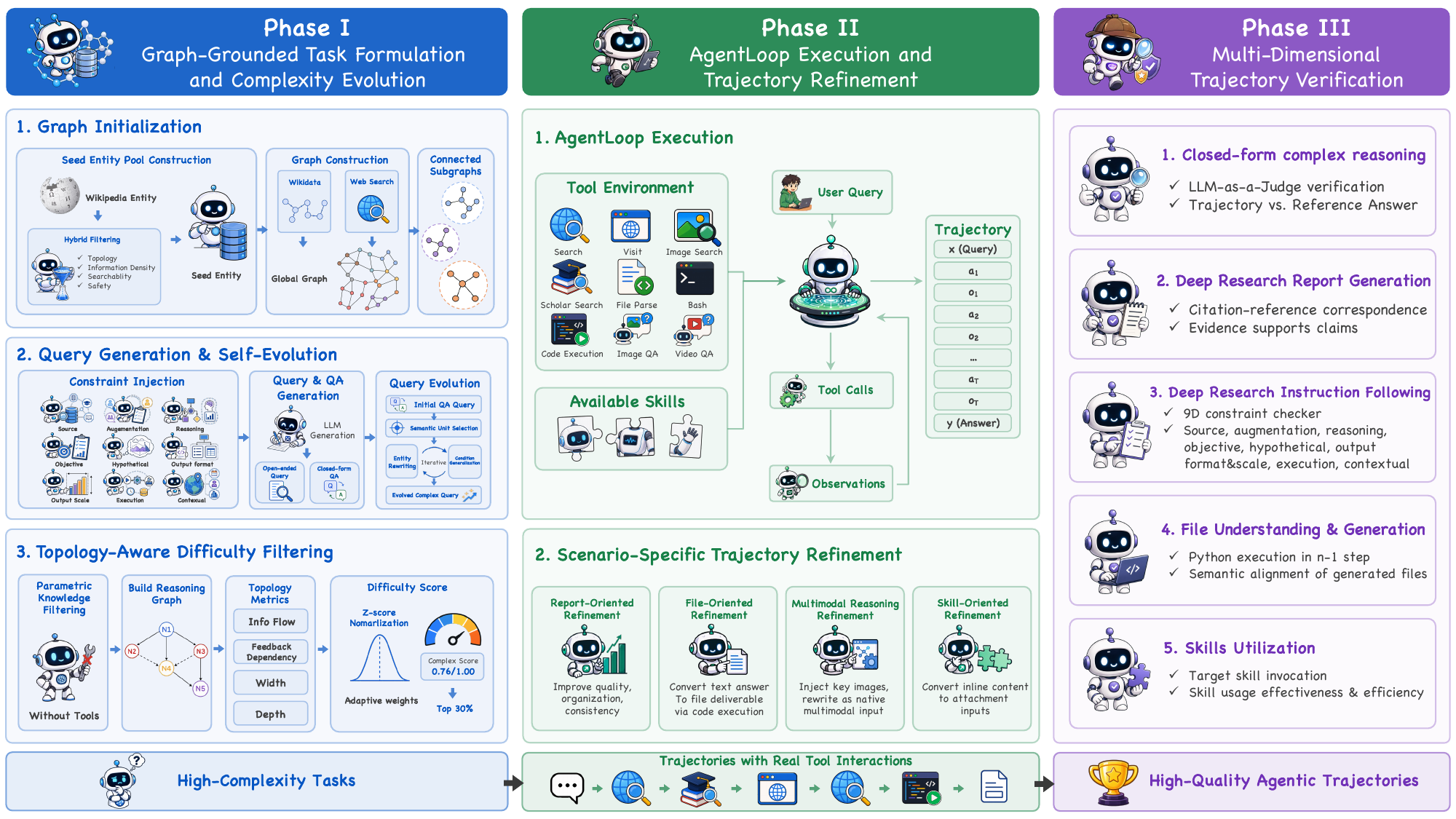}
    \caption{Overview of the Agentic Data Synthesis Framework.}
    \label{fig:framework}
\end{figure*}

Motivated by these observations, S1-DeepResearch aims to construct a unified data foundation that extends deep research agents beyond search-centric task paradigms. The proposed system synthesizes complex and verifiable agentic trajectories across multiple deep research capabilities through three major stages, as illustrated in Figure~\ref{fig:framework}.

\textbf{Phase I: Graph-Grounded Task Formulation and Complexity Evolution.} This stage constructs complex queries in a top-down manner by leveraging connected subgraphs from knowledge graphs as structured knowledge backbones. During task generation, explicit research constraints, such as required information sources, output formats, and quantity limitations, are injected into the subgraph in advance. To prevent models from solving tasks solely based on their parametric knowledge and bypassing tool usage (Tool-use Bypass), we introduce a parametric-knowledge-based filtering mechanism, together with graph-topology-based complexity filtering, to select challenging tasks that require multi-step exploration.

\textbf{Phase II: AgentLoop Execution and Trajectory Refinement.}
After structured task generation and difficulty filtering, this stage converts the selected tasks into executable tool-interactive trajectories through AgentLoop within a sandbox environment. During rollout, the agent iteratively interacts with nine categories of atomic tools, including search, webpage browsing, code execution, and multimodal question answering, producing multi-step trajectories driven by environmental feedback. The rollout process also covers skill-aware trajectory construction, enabling the collected data to include both general tool-use behaviors and specialized skill-use behaviors. Based on the collected trajectories, we then apply scenario-specific refinement to improve the coverage and quality of the resulting training data. This refinement reshapes selected trajectories into more realistic task forms, such as native multimodal queries, uploaded-file inputs, and executable artifact delivery, while also enhancing final deliverables with stricter quality requirements, such as deep research reports.

\textbf{Phase III: Multi-Dimensional Trajectory Verification.}
The final stage performs multi-dimensional automated verification to reduce hallucinations and ensure data correctness. We design specialized verifiers for the five capability dimensions and apply strict conditional filtering strategies. For example, in deep research scenarios, the verifier examines whether in-text citations are correctly aligned with the reference list and whether the cited evidence provides substantial support for the corresponding claims. Finally, only trajectories that satisfy all requirements regarding action sequences, multi-dimensional constraints, and academic standards are retained as training data.

\subsection{Phase I: Graph-Grounded Task Formulation and Complexity Evolution}

\subsubsection{Graph Initialization}

\textbf{Seed Entity Pool Construction.} 
To ensure broad knowledge coverage across interdisciplinary domains, we first initialize a basic entity pool from Wikipedia Entities and construct a diverse candidate entity set through hierarchical sampling. To ensure that the generated research instructions can effectively trigger multi-hop reasoning while avoiding unsolvable retrieval scenarios, we introduce a Hybrid Filtering Mechanism to identify high-quality seed entities.

Specifically, the filtering pipeline consists of four progressive stages. First, we perform topology- and popularity-based filtering by considering the number of Sitelinks and the scale of 2-hop neighbors. This removes both semantically isolated long-tail entities and overly generic high-frequency entities, thereby selecting entities with an appropriate exploration space. Second, we apply low-information-density filtering to remove temporal entities and purely numerical entities, ensuring that the remaining nodes serve as meaningful semantic anchors rather than simple relational modifiers. Third, we conduct searchability verification by requiring each candidate entity to retrieve a sufficient number of valid web results, ensuring adequate digital footprints and knowledge consensus for subsequent cross-document exploration. Finally, we apply safety filtering to exclude entities involving harmful, sensitive, or highly controversial content, mitigating potential risks of biased generation, unsafe responses, and unstable retrieval behavior. Through this hybrid filtering pipeline, we obtain a seed entity pool with high knowledge density, clear semantic grounding, and strong verifiability.

\textbf{Subgraph Construction.}
For each entity in the seed entity pool, we adopt a dual-path expansion strategy to construct a corresponding directed acyclic graph (DAG). The first path leverages structured knowledge from Wikidata, where factual triples associated with the seed entity are expanded through multi-hop relation traversal to gradually establish the topological structure among entities. The second path relies on open-domain search engines to dynamically supplement external knowledge from the web and enrich semantic connections beyond structured knowledge bases.

For open-domain retrieval results, we further introduce a multimodal information parsing mechanism. At the textual level, we perform entity recognition, relation extraction, and event summarization over webpage content to capture deep semantic associations. At the visual level, we parse visual entities from embedded images, charts, and scene information, align them with textual semantics, and incorporate the extracted visual concepts as heterogeneous nodes into the DAG topology. By jointly modeling textual semantic relations and visual entity associations, we construct a unified heterogeneous knowledge network, which provides structured support for subsequent data synthesis involving multimodal reasoning, multi-hop inference, and complex research task generation.

After graph expansion, the resulting global knowledge network is usually large-scale and sparsely connected. Therefore, we further apply community detection algorithms to partition the DAG into multiple connected subgraphs with dense internal connections and high semantic coherence. Compared with the original global graph, these connected subgraphs better represent high-order knowledge interaction regions and potential reasoning paths within specific semantic contexts, serving as structured foundations for complex task generation and trajectory rollout.

\subsubsection{Query Generation and Self-Evolution}

\textbf{Constraint Injection.}
For open-ended deep research tasks, conventional question generation methods often lack explicit constraints on research scope and reasoning objectives, which may lead to loosely structured tasks, objective drift, or shallow information aggregation.
To improve the complexity and executability of generated tasks, we introduce a \textbf{Pre-generation Constraint Injection} mechanism before complex query synthesis. Given the domain attributes, relational topology, and semantic dependencies derived from the graph structure, this mechanism imposes structured constraints on the task generation process, explicitly defining the exploration space, reasoning paths, and execution boundaries of the generated tasks. Specifically, we first model potential research objectives based on the entity type distribution, cross-domain relationships, and structural complexity of the connected subgraph.
Then, multi-dimensional constraints are dynamically sampled from a predefined constraint space and integrated with the graph structure to form a unified task conditioning context.

To enhance the controllability and reasoning requirements of open-ended research tasks, we construct a comprehensive constraint space consisting of nine dimensions:
\textit{Source Constraints},
\textit{Argumentation Constraints},
\textit{Reasoning Constraints},
\textit{Objective Constraints},
\textit{Hypothetical Constraints},
\textit{Output Format Constraints},
\textit{Output Scale Constraints},
\textit{Execution Constraints},
and \textit{Contextual Constraints}.
Detailed definitions of each constraint dimension are provided in Appendix~\ref{app:constraints_space}.

\textbf{Query/QA Generation.}
After connected subgraph extraction and constraint injection, we leverage large language models to generate corresponding natural language tasks. Given a connected subgraph and a set of constraints, we adopt two generation paradigms according to different task objectives: open-ended research task generation and closed-form question-answer generation.

\begin{itemize}
    \item \textbf{Open-Ended Query Generation.}
    Given the connected subgraph and sampled constraints as the conditioning context, we generate research-oriented queries requiring complex multi-stage reasoning. During generation, the model is encouraged to fully utilize heterogeneous information contained in the subgraph, including entity attributes, structural relations, and visual semantics. Meanwhile, the injected constraints are naturally incorporated into the task description, ensuring that the generated queries have explicit research boundaries, executable objectives, and traceable factual foundations.
    \item \textbf{Closed-Form QA Generation.}
    For question-answering scenarios with deterministic answers, we jointly generate questions and corresponding answers based on the connected subgraph. Unlike open-ended task generation, this process emphasizes factual consistency and alignment between generated QA pairs and the underlying graph structure. Specifically, factual information contained in the answers should be traceable to entity nodes, relational edges, or associated attributes in the subgraph, ensuring clear evidence sources and reliable verification criteria.
\end{itemize}

\textbf{Semantics-Driven Query Evolution.}
For closed-form QA tasks, we introduce an iterative \textbf{Query Evolution} process to increase reasoning complexity while preserving answer accessibility. The key idea is to gradually weaken the direct association between explicit surface clues and the target answer through multi-round query rewriting. At each iteration, we randomly select semantic units from the current query and perform a combination of two operations:
(1) \textbf{Entity Semantic Rewriting}: For entity-related information, we retrieve background descriptions and relevant attributes through search engines, and replace explicit entity mentions with context-dependent descriptions, such as functional roles, historical behaviors, or relational expressions. This reduces the dependence on direct entity matching and encourages deeper reasoning.
(2) \textbf{Condition Semantic Generalization}: For constraints involving time, location, numerical values, or events, we transform precise expressions into higher-level or more abstract semantic representations, such as temporal interval expansion, spatial region generalization, or event-level abstraction. This reduces directly searchable anchor information and increases the requirement for multi-hop reasoning.

\subsubsection{Topology-Aware Difficulty Filtering}

\textbf{Parametric Knowledge Filtering.}
To identify tasks that require external exploration and complex reasoning, we first evaluate generated Query or QA pairs under a tool-free setting, where all external tools, including retrieval and document parsing modules, are disabled.
The generated queries are directly provided to the base language model.
If the model can solve the task solely relying on its parametric knowledge or complete the reasoning process without external information, the task is regarded as a low-complexity sample and removed from the dataset.

\textbf{Topology-Aware Difficulty Estimation.}
For closed-form QA tasks, after multiple rounds of Query Evolution,
the entity expressions, constraints, and reasoning paths in the final query
may substantially deviate from the initial connected subgraph.
Therefore, directly estimating task difficulty based on the original
knowledge subgraph can no longer accurately reflect the actual reasoning
complexity.

To address this issue, we construct a task-specific
\textbf{Reasoning Graph} for each evolved QA sample.
Specifically, we leverage LLMs to analyze the expected reasoning process
and convert it into a structured graph $G=(V,E)$,
where nodes represent key entities, constraints, intermediate conclusions,
and final answers, while edges represent dependency and reasoning relations.
Each reasoning graph is further categorized into different structural
patterns, including sequential, convergent, divergent, comparative,
and graph-based reasoning.

Given the constructed reasoning graph, we quantify its structural complexity
from four topology-aware dimensions.

\textbf{Information Flow Complexity}
measures the degree of information aggregation and propagation through
intermediate reasoning nodes.
Inspired by information flow metrics in software structure analysis~\citep{kontogiannis1996pattern},
we define:

\[
\mathcal{C}_{info}(G)
=
\sum_{v \in V\setminus\{v_s,v_t\}}
(d^{-}(v)d^{+}(v))^2 ,
\]

where $v_s$ and $v_t$ denote the source and target nodes,
and $d^{-}(v)$ and $d^{+}(v)$ denote the in-degree and out-degree of node $v$.
A higher value indicates stronger cross-path information dependency.

\textbf{Feedback Dependency Complexity}
measures cyclic dependencies and non-linear reasoning structures in the
reasoning graph.
Inspired by Kahn's topological sorting algorithm~\citep{kahn1962topological}, which
iteratively removes zero in-degree nodes to identify acyclic structures,
and subsequent topological peeling methods for complex network
analysis~\citep{chen2002experimental}, we adopt a heuristic topological peeling strategy
to approximate the minimum feedback edge set (MFES) of the reasoning graph.

Let $\hat{\mathcal{E}}_{fb}$ denote the approximated feedback edge set obtained
through the peeling process. We define:

\[
\mathcal{C}_{fb}(G)
=
|\hat{\mathcal{E}}_{fb}|,
\]

where each edge in $\hat{\mathcal{E}}_{fb}$ represents a dependency relation
that participates in cyclic reasoning structures.
A larger $\mathcal{C}_{fb}(G)$ indicates stronger iterative dependencies
among intermediate reasoning states, requiring the model to maintain
consistency across multiple reasoning paths.

After removing the approximated feedback edges, we transform the reasoning graph into a DAG
$\tilde{G}=(V,\tilde{E})$, where
$\tilde{E}=E\setminus\hat{\mathcal{E}}_{fb}$.

After removing feedback edges, we transform the reasoning graph into a DAG
$\tilde{G}=(V,\tilde{E})$ and further analyze its hierarchical structure.

\textbf{Width Complexity}
measures the maximum number of parallel reasoning branches.
Given the layer partition
$\{\mathcal{L}_1,\mathcal{L}_2,\dots,\mathcal{L}_K\}$
obtained by BFS traversal, we define:

\[
\mathcal{C}_{width}(\tilde{G})
=
\max_{1\leq k\leq K}
|\mathcal{L}_k|,
\]

where $\mathcal{L}_k$ denotes the node set at the $k$-th reasoning layer.

\textbf{Reasoning Depth Complexity}
measures the longest dependency chain from the source node to the target node:

\[
\mathcal{C}_{depth}(\tilde{G})
=
\max_{\pi\in\mathcal{P}(v_s,v_t)}
|\pi|
\]

where $\mathcal{P}(v_s,v_t)$ denotes all valid reasoning paths from $v_s$ to $v_t$.

Since reasoning graphs differ significantly in scale and density,
the raw values of different topology-aware metrics are not directly comparable.
Therefore, we normalize each metric using Z-score normalization.

Let $\phi_m$ denote the raw value corresponding to metric type $m$:

\[
\phi_m \in
\{
\mathcal{C}_{info}(G),
\mathcal{C}_{fb}(G),
\mathcal{C}_{width}(\tilde{G}),
\mathcal{C}_{depth}(\tilde{G})
\},
\quad
m\in\mathcal{M},
\]

where $\mathcal{M}=\{info,fb,width,depth\}$.
The normalized score is computed as:

\[
z_m
=
\frac{\phi_m-\mu_m}{\sigma_m},
\]

where $\mu_m$ and $\sigma_m$ denote the mean and standard deviation
of metric type $m$ across all reasoning graphs:

\[
\mu_m
=
\frac{1}{N}
\sum_{i=1}^{N}
\phi_m^{(i)},
\]

\[
\sigma_m
=
\sqrt{
\frac{1}{N}
\sum_{i=1}^{N}
(\phi_m^{(i)}-\mu_m)^2
}.
\]

The final topology-aware difficulty score is obtained by aggregating
all normalized metrics:

\[
\mathcal{D}
=
\sum_{m\in\mathcal{M}}
w_m z_m ,
\]

where $w_m$ denotes the weight assigned to each topology-aware metric.
Finally, we rank all generated tasks according to $\mathcal{D}$ and retain
only the top 30\% most complex samples within each batch for trajectory
generation in Phase II.

\subsection{Phase II: AgentLoop Execution and Trajectory Refinement}

\subsubsection{End-to-End Rollout via AgentLoop}

\textbf{Tool Environment.}
After task generation and difficulty filtering, we use AgentLoop as an execution harness to synthesize complete tool-interactive trajectories in an executable sandbox environment. AgentLoop is equipped with nine categories of atomic tools: \textbf{web search},
\textbf{web visit},
\textbf{image search},
\textbf{academic search},
\textbf{file parsing},
\textbf{code execution},
\textbf{bash},
\textbf{image question answering},
and \textbf{video question answering}. These tools collectively cover the major operations involved in deep research workflows, including open-web and scholarly information acquisition, source-level evidence inspection, heterogeneous file understanding, deterministic computation and programmatic verification, multimodal evidence analysis, and shell-level interaction with the sandbox environment. Detailed tool descriptions and schemas are provided in Appendix \ref{app:tool_enviroment_details}.

\textbf{Tool-Interactive Trajectory Construction.}
Given a filtered task, AgentLoop places the agent in the sandbox and drives an end-to-end interaction process between the model and the executable environment. At each step, the agent observes the current context, decides the next reasoning or tool-use action, invokes the corresponding tool when needed, and receives environmental feedback such as retrieved webpages, parsed file contents, execution results, visual evidence, or error messages. This closed-loop process converts a task instruction into a complete agentic trajectory with explicit action-observation transitions:
$$
\tau = (x, a_1, o_1, a_2, o_2, \ldots, a_T, o_T, y),
$$
where $x$ denotes the task instruction, $a_t$ denotes the model action or tool call at step $t$, $o_t$ denotes the corresponding environmental observation, and $y$ denotes the final response. The rollout result is regarded as a candidate training trajectory. Assistant-side reasoning steps, tool calls, and final answers are retained as potential supervision signals, whereas tool observations are treated as external environment outputs.

\textbf{Skill-Aware Rollout.}
AgentLoop also covers skill-aware trajectory construction. In this setting, the sandbox mounts the target skill together with several distractor skills, while exposing only lightweight metadata and the path to each \texttt{SKILL.md} file. The agent must inspect the relevant documentation through atomic tools such as \texttt{bash} before triggering the underlying scripts according to the documented usage. This progressive loading mechanism allows the rollout to capture skill selection, documentation grounding, and script-level execution within the same tool-interactive trajectory.

\subsubsection{Scenario-Specific Trajectory Refinement}

AgentLoop converts filtered tasks into executable tool-interactive trajectories, which serve as the initial training data for agentic behavior learning. We further apply scenario-specific trajectory refinement as a post-processing stage to improve the coverage and quality of the collected data across different deep research scenarios. Specifically, it reshapes selected trajectories into more realistic task forms and enhances outputs that require stricter deliverable standards.

\textbf{Report-Oriented Refinement.}
For deep research report generation, refinement focuses on improving the quality and consistency of the final deliverable. We first assess the relevance of the collected evidence, then derive a report outline and expected evaluation criteria from the user query, task constraints, and evidence. The final report is rewritten under these signals to improve evidence coverage, organization, logical consistency, and format compliance. We further refresh the last \texttt{Think} step to keep the model's final reasoning state aligned with the delivered report.

\textbf{File-Oriented Refinement.}
For file understanding and generation tasks, refinement converts text-based outputs into file-based deliverables. We first inject explicit file requirements into the original query, such as the target format, schema, layout, or content organization. We then transform the original textual answer into a file-generation step using \texttt{execute\_code}. The final response summarizes the generated file and its completion status, turning the trajectory from textual answering into executable artifact delivery.

\textbf{Multimodal Reasoning Refinement.}
For multimodal long-horizon reasoning, refinement transforms text-only queries into native multimodal inputs. When the original trajectory relies on a key image, we insert this image into the query and rewrite the task around its visual clues. When the trajectory depends on a key entity, we retrieve a representative image containing that entity and combine it with the original query. This preserves the key visual context involved in the original reasoning process while presenting the task in a more realistic multimodal input form.

\textbf{Skill-Oriented Refinement.}
For skill-based execution tasks, refinement makes the input form closer to real user interactions. For queries that directly include code snippets, configuration blocks, or structured contents such as JSON, CSV, YAML, or scripts, we sample a subset and convert the inline content into attachment inputs. The query is then rewritten to refer to the uploaded file, requiring the model to identify the file type and task intent, inspect the relevant skill documentation, execute the appropriate tool, and summarize the result.

\subsection{Phase III: Multi-Dimensional Trajectory Verification}

The objective of this stage is to impose systematic quality control and consistency verification at the trajectory level, thereby reducing hallucination accumulation and improving the verifiability and traceability of generated agent behaviors across complex reasoning, deep research, and tool-use scenarios. Instead of relying solely on final-answer evaluation, we introduce a multi-dimensional verification framework that examines intermediate trajectories and performs task-specific consistency checking.

For \textbf{closed-form complex reasoning tasks}, we adopt an LLM-as-a-Judge based discriminative verification framework with reference-answer alignment. Specifically, the verifier performs structured comparisons between the generated reasoning trajectory and the reference answer, evaluating logical correctness, reasoning completeness, and the validity of intermediate derivations. Compared with answer-only matching, this trajectory-level evaluation provides finer-grained supervision and better captures error propagation during multi-step reasoning.

For \textbf{deep research report generation tasks}, verification focuses on factual consistency and adherence to academic standards.
We construct a citation-aware verifier to examine the correspondence between in-text citations and reference entries, ensuring the completeness and traceability of citation chains.
Furthermore, we evaluate whether the cited evidence semantically supports the corresponding claims, mitigating citation hallucinations where references appear formally valid but provide insufficient or irrelevant evidence.

For \textbf{deep research instruction-following tasks}, we introduce a \textbf{nine-dimensional constraint checker} to perform strict trajectory-level validation.
The verifier jointly evaluates whether the generated trajectory satisfies predefined constraints across source, argumentation, reasoning, objective, hypothetical, output format, output scale, execution, and contextual dimensions.
These constraints cover information acquisition boundaries, reasoning assumptions, evidence organization, tool-use behaviors, and output requirements.
Any violation of these constraints causes the trajectory to be filtered out, ensuring strict alignment with complex user instructions.

For \textbf{file understanding and generation tasks}, we focus on the consistency between agent trajectories and executable code operations.
In particular, for multi-step generation workflows, we require the trajectory to include Python execution within the final execution stages, ensuring that intermediate reasoning and data processing steps are supported by verifiable computation.
In addition, we evaluate the semantic alignment between generated files and the original task requirements, verifying whether the outputs satisfy constraints regarding structure, information coverage, and presentation format.

For \textbf{skill-use tasks}, verification mainly focuses on skill activation effectiveness and utilization efficiency.
We track the invocation paths of Target Skills to determine whether they are correctly activated and involved in the decision-making process.
Moreover, we leverage LLM-as-a-Judge to evaluate the usage of external knowledge resources and executable skill modules associated with each skill, measuring the contribution of each skill module to the final trajectory.
This verification process helps identify redundant, inactive, or incorrectly invoked skills, providing feedback for subsequent skill library optimization and routing strategy improvement.

\subsection{Dataset Statistics and Capability Distribution}

After the three-stage pipeline of task formulation, trajectory rollout, and multi-dimensional verification, we obtain a large-scale agent trajectory dataset covering the full Deep Research workflow. Unlike existing agent datasets that are predominantly search-centric, S1-DeepResearch encompasses a broader spectrum of task types, including knowledge synthesis, complex reasoning, and planning and decision making. To characterize the resulting dataset, we analyze it from two complementary perspectives: 
(1) the dataset-level differences between S1-DeepResearch and existing agent trajectory datasets, and 
(2) the category-level differences within S1-DeepResearch, where different statistical profiles reveal distinct capability demands across Deep Research tasks.

\begin{table*}[t]
\centering
\small
\caption{
Comparison of S1-DeepResearch and existing open-source deep research datasets.
All length-related metrics are measured in tokens.
Tool Calls and Tool Types are averaged over trajectories.
Traj Think, Step Think, and Final Think denote thinking-token statistics at different granularities.
Tool Pool refers to the total number of tools available to the agent.
}
\label{tab:dataset_comparison}
\resizebox{\textwidth}{!}{
\begin{tabular}{lrrrrrrrrr}
\toprule

Dataset &
Samples &
\makecell{Tool\\Calls} &
\makecell{Total\\Len.} &
\makecell{Traj\\Think} &
\makecell{Step\\Think} &
\makecell{Answer\\Len.} &
\makecell{Final\\Think} &
\makecell{Tool\\Types} &
\makecell{Tool\\Pool} \\

\midrule

REDSearcher (Text) & 10001 & 64.1 & 59890 & 10149 & 156 & 234 & 349 & 2.12 & 5 \\
REDSearcher (MM) & 5816 & 12.2 & 12830 & 3582 & 272 & 236 & 753 & 3.27 & 6 \\
OpenSeeker (All) & 11677 & 46.1 & 73835 & 19181 & 408 & 349 & 579 & 1.94 & 2 \\
OpenSeeker (Correct) & 4949 & 27.2 & 51165 & 13063 & 466 & 357 & 623 & 1.89 & 2 \\
OpenSeeker (Incorrect) & 6728 & 60.1 & 90511 & 23681 & 389 & 342 & 547 & 1.97 & 2 \\
OpenResearcher & 97630 & 52.6 & 55090 & 5628 & 105 & 214 & 617 & 2.80 & 4 \\
\midrule
S1-DeepResearch & 49299 & 9.7 & 20431 & 2524 & 273 & 1739 & 552 & 1.92 & 9 \\

\bottomrule
\end{tabular}
}
\end{table*}

\begin{figure}[t]
    \centering

    \begin{minipage}[t]{0.48\linewidth}
        \centering
        \includegraphics[width=\linewidth, height=0.28\textheight,keepaspectratio]{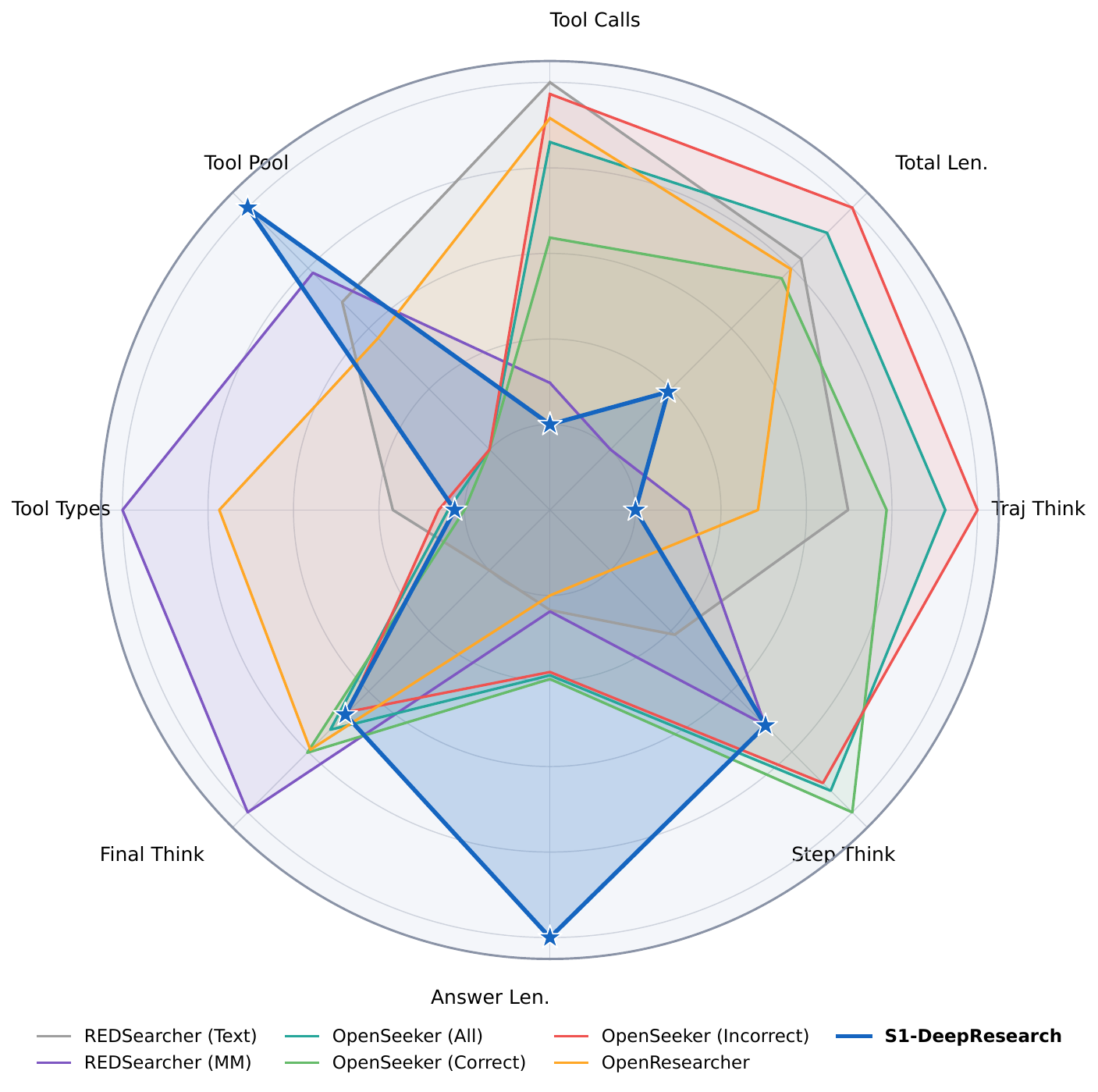}
        \captionof{figure}{Comparison of Agentic Trajectories. For visualization, all metrics are log-scaled and min-max normalized to the range [0.2, 1.0].}
        \label{fig:comparison_agentic_trajectories}
    \end{minipage}
    \hfill
    \begin{minipage}[t]{0.48\linewidth}
        \centering
        \includegraphics[height=0.28\textheight,keepaspectratio]{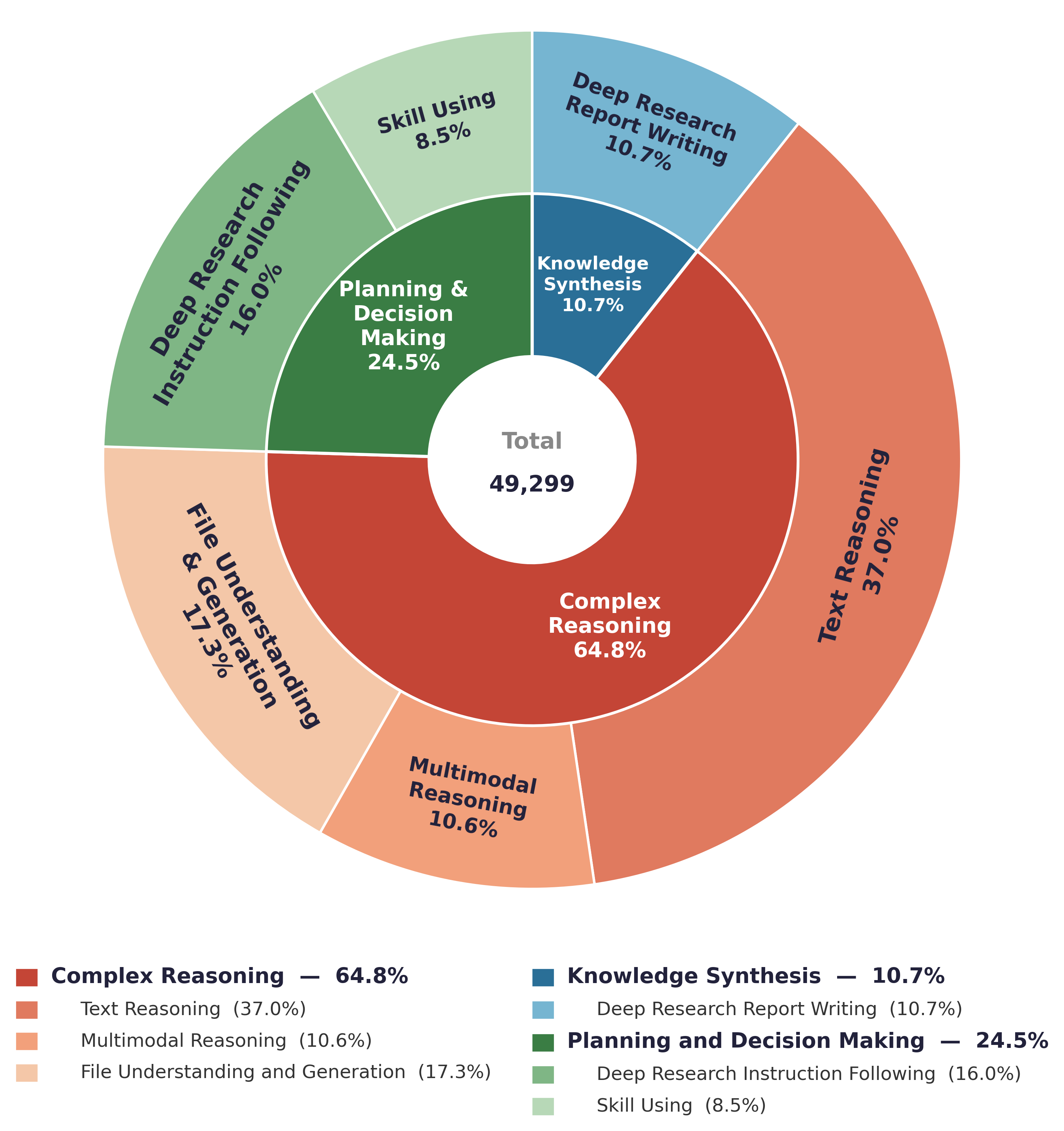}
        \captionof{figure}{Capability Composition of S1-DeepResearch.}
        \label{fig:sunburst_dataset}
    \end{minipage}
\end{figure}

\begin{figure}[t]
    \centering

    \begin{minipage}[t]{0.48\linewidth}
        \centering
        \includegraphics[width=\linewidth, height=0.22\textheight,keepaspectratio]{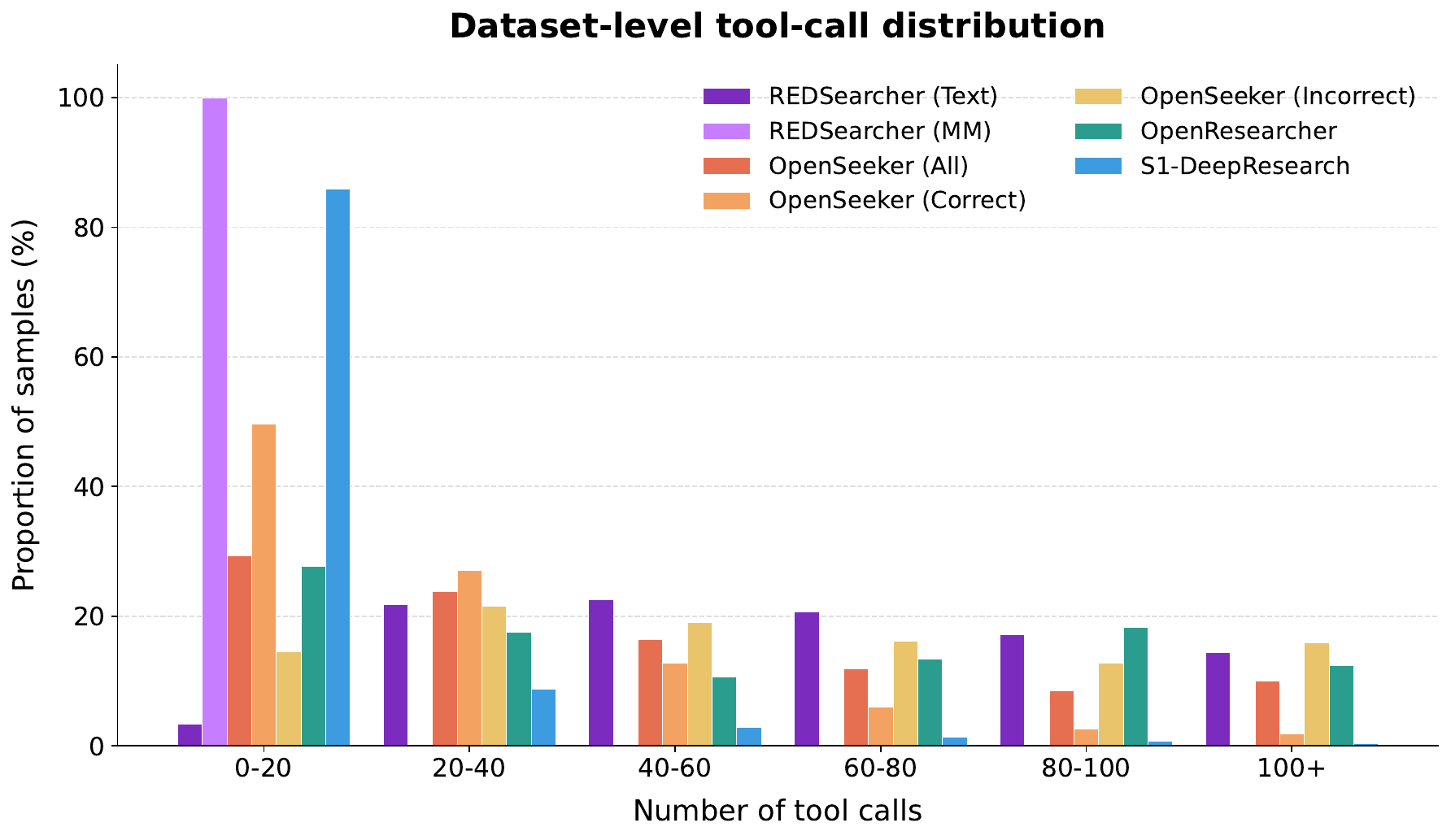}
        \captionof{figure}{
        Dataset-level tool-call distribution across agentic trajectory datasets.
        }
        \label{fig:dataset_tool_call_distribution}
    \end{minipage}
    \hfill
    \begin{minipage}[t]{0.48\linewidth}
        \centering
        \includegraphics[width=\linewidth, height=0.22\textheight,keepaspectratio]{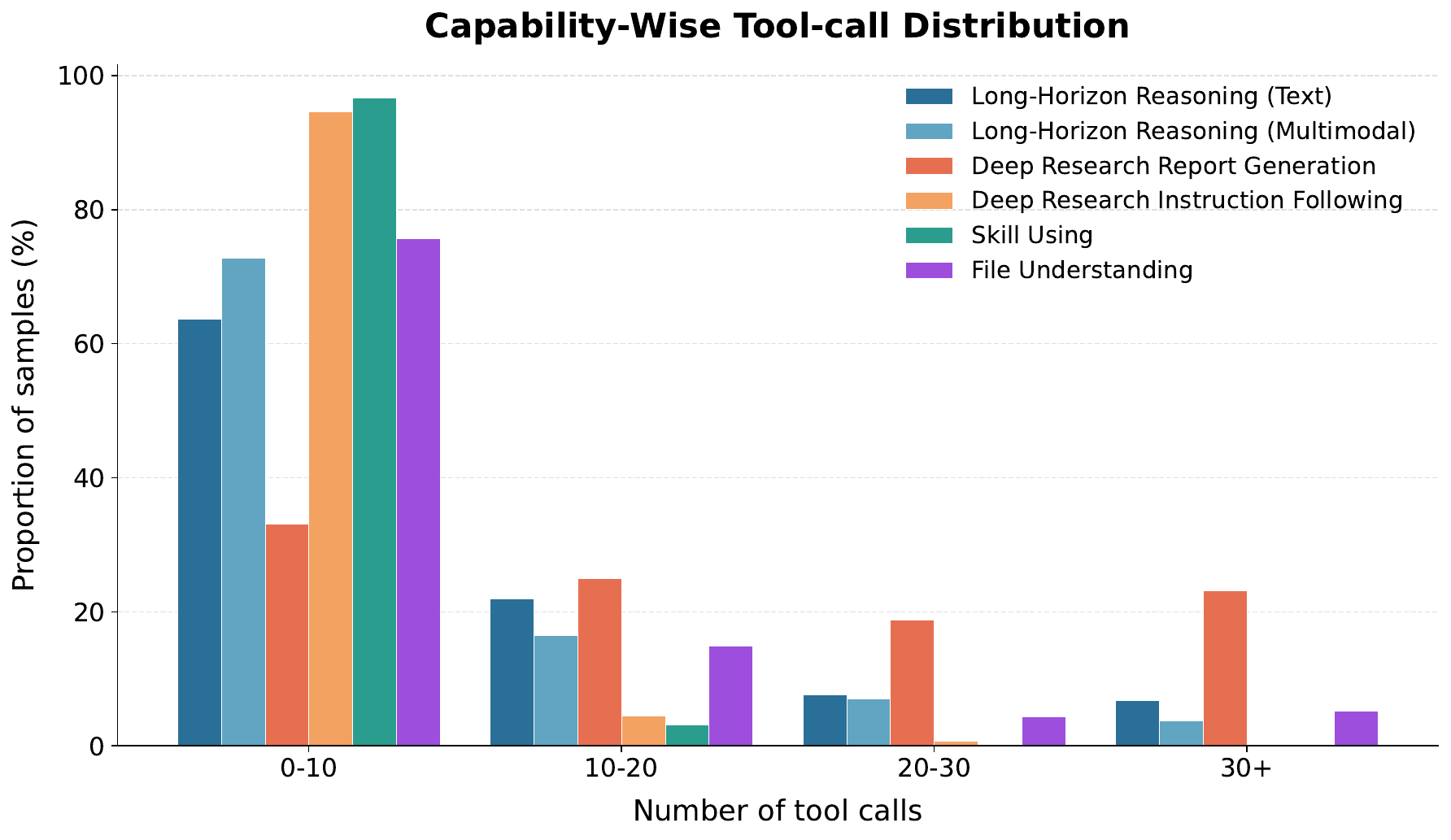}
        \captionof{figure}{
        Capability-wise tool-call distribution within S1-DeepResearch.
        }
        \label{fig:s1_tool_call_distribution}
    \end{minipage}
\end{figure}

\textbf{Dataset-Level Trajectory Analysis.}
We first characterize the trajectory profiles of different agent datasets and analyze what these profiles reveal about their data construction objectives and task orientations. As shown in Table~\ref{tab:dataset_comparison}, Figure~\ref{fig:comparison_agentic_trajectories}, and Figure~\ref{fig:dataset_tool_call_distribution}, existing datasets such as REDSearcher, OpenSeeker, and OpenResearcher exhibit a search-centric trajectory pattern.
Their trajectories typically contain many tool invocations and long intermediate reasoning traces, while the final responses remain relatively short.
This suggests that these datasets mainly emphasize the process of information acquisition, including query formulation, source inspection, evidence verification, and candidate answer selection. Such data is valuable for training agents to explore and verify information, but it provides limited coverage of several critical capabilities required by Deep Research. On the input side, existing datasets rarely involve complex research instructions with explicit constraints on sources, arguments, reasoning procedures, output formats, or execution conditions. On the output side, they insufficiently cover downstream research completion stages, where the agent must organize evidence, construct arguments, synthesize findings, and produce usable research outputs.

S1-DeepResearch presents a different trajectory profile. It uses fewer tool calls on average than most search-centric datasets, but produces substantially longer and more structured final responses. This contrast indicates that the key distinction of S1-DeepResearch is not simply the length or interaction frequency of trajectories, but the shift of trajectory design from \emph{search-intensive answer finding} to \emph{research-oriented task completion}. In S1-DeepResearch, retrieval and tool use serve as intermediate steps for grounding the response, while the final objective is to complete diverse research-oriented tasks under explicit requirements. These tasks include synthesizing evidence into long-form reports, following complex research instructions, generating document artifacts, and executing specialized skills. As a result, S1-DeepResearch covers a more complete task execution process, spanning information acquisition, constraint interpretation, decisions about when the collected evidence is sufficient, and research outcome construction.

The trajectory-level contrast further highlights the limitation of simply increasing search depth. A closer examination of OpenSeeker shows that unsuccessful trajectories can involve even more tool calls and longer reasoning chains than successful ones, suggesting that longer exploration does not necessarily lead to better task completion in complex open-domain settings. When an agent repeatedly revisits candidate answers or remains trapped in verification loops, additional tool calls may contribute little to the final outcome. S1-DeepResearch therefore aims to balance evidence acquisition, constraint satisfaction, and output construction. The resulting trajectories preserve sufficient tool interaction for evidence grounding, while emphasizing when to stop searching, how to satisfy task-specific constraints, and how to organize retrieved information into high-quality research outcomes. In other words, the core challenge of Deep Research is not to construct longer search trajectories, but to transform sufficiently comprehensive retrieved information into reliable, well-structured, and usable research results in a timely and effective manner.

\begin{table*}[t]

\centering

\small

\caption{Detailed Statistics by Data Category. All length-related metrics are measured in tokens. Tool Calls and Tool Types are averaged over trajectories. Traj Think, Step Think, and Final Think denote thinking-token statistics at different granularities. Tool Pool refers to the total number of tools available to the model.
}

\label{tab:s1-deepresearch_data_statistics}

\resizebox{\textwidth}{!}{

\begin{tabular}{lrrrrrrrrr}

\toprule

Dataset & Samples & \makecell{Tool\\Calls} &

\makecell{Total\\Len.} &

\makecell{Traj\\Think} &

\makecell{Step\\Think} &

\makecell{Answer\\Len.} &

\makecell{Final\\Think} &

\makecell{Tool\\Types} &

\makecell{Tool\\Pool}\\
\midrule

Long-Horizon Reasoning (Text) & 18246 & 10.7 & 18830 & 2916 & 249 & 879 & 398 & 1.93 & 4 \\
Long-Horizon Reasoning (Multimodal) & 5203 & 8.8 & 21258 & 1167 & 127 & 363 & 214 & 2.96 & 8 \\
Deep Research Report Generation & 5258 & 23.5 & 51625 & 3904 & 178 & 8921 & 1241 & 1.29 & 4 \\
Deep Research Instruction Following & 7877 & 3.9 & 16213 & 2050 & 422 & 1626 & 786 & 1.42 & 6 \\
Skill Using & 4208 & 3.7 & 12706 & 2255 & 450 & 1402 & 646 & 1.94 & 8 \\
File Understanding \& Generation & 8507 & 8.0 & 11803 & 2234 & 245 & 257 & 401 & 2.10 & 6 \\
\midrule
Overall & 49299 & 9.7 & 20431 & 2524 & 273 & 1739 & 552 & 1.92 & 9 \\
\bottomrule
\end{tabular}
}
\end{table*}

\textbf{Capability-Wise Trajectory Analysis.}
We examine the capability-wise statistics of S1-DeepResearch to characterize the trajectory profiles associated with different data categories.
Figure~\ref{fig:sunburst_dataset}, Table~\ref{tab:s1-deepresearch_data_statistics}, and Figure~\ref{fig:s1_tool_call_distribution} show clear variations across tool-call frequency, reasoning-token distribution, answer length, and tool-type usage.
Specifically, the data exhibits three representative trajectory patterns.
(1) \emph{Output-intensive trajectories.} Deep research report generation has the highest tool-call frequency, the longest total trajectories, and the longest final answers, indicating that this category combines substantial external interaction with extended final-output construction.
(2) \emph{Decision-intensive trajectories.} Deep research instruction following and skill using require relatively few tool calls but have the longest step-level thinking lengths, suggesting that more reasoning effort is concentrated before each action, especially for constraint interpretation, action selection, and execution control.
(3) \emph{Tool-heterogeneous trajectories.} Multimodal reasoning activates the largest number of tool types per trajectory, while file understanding \& generation combines non-trivial tool usage with short final answers, indicating that a considerable portion of the task process is reflected in intermediate perception, parsing, execution, or artifact-generation steps rather than in final textual generation.

Overall, the capability-wise statistics show that S1-DeepResearch preserves search-centric long-horizon reasoning trajectories while extending to several additional Deep Research scenarios. These additional categories exhibit trajectory patterns different from search-centric exploration: report generation emphasizes long-form output construction, instruction following and skill using emphasize constraint-aware decision making, and file-centric or multimodal tasks emphasize heterogeneous tool coordination. Together, these trajectory patterns complement search-centered data and provide broader coverage of the data forms required in realistic Deep Research scenarios.

\section{Training}

We instantiate S1-DeepResearch from Qwen3-32B~\citep{yang2025qwen3}, a reasoning-oriented open-weight model that provides a potential initialization for instruction following, long-horizon reasoning, and tool-use adaptation. We then perform supervised fine-tuning on curated agentic trajectories, where each trajectory contains a user task, intermediate model actions, environment observations, and a final response. Formally, a trajectory is represented as
\[
\tau = (x, a_1, o_1, a_2, o_2, \ldots, a_T, y),
\]
where \(x\) denotes the user task, \(a_t\) denotes the model-generated reasoning step or tool action at step \(t\), \(o_t\) denotes the environment observation returned after \(a_t\), and \(y\) denotes the final response. During training, supervision is applied to assistant-side actions and final answers, while environment observations are treated as external outputs and excluded from the loss. Accordingly, we optimize the following supervised trajectory imitation objective:
\[
\mathcal{L}
=
-\sum_{t=1}^{T}
\log p_{\theta}\!\left(a_t \mid x, a_{<t}, o_{<t}\right)
-
\log p_{\theta}\!\left(y \mid x, a_{\leq T}, o_{\leq T}\right).
\]
This objective encourages the model to imitate high-quality research trajectories by learning both how to select intermediate actions conditioned on accumulated observations and how to synthesize the final response after sufficient evidence has been gathered.

\section{Experiments}
We conduct extensive experiments to evaluate S1-DeepResearch across diverse deep research scenarios. The evaluation includes comparisons with existing models and systems, analysis of test-time exploration behavior.

\subsection{Experimental Setup}

\subsubsection{Benchmarks}

We evaluate S1-DeepResearch with a comprehensive benchmark suite covering five core capabilities required by deep research agents:
\textbf{long-horizon complex reasoning},
\textbf{deep research report generation},
\textbf{deep research instruction following},
\textbf{file understanding and generation},
and \textbf{dynamic skill utilization}.
The evaluation suite consists of established public benchmarks for standardized comparison and in-house benchmarks designed to assess more open-ended research scenarios.

\begin{itemize}

    \item \textbf{Long-horizon complex reasoning.}
    We evaluate the model's ability to perform multi-step exploration, information seeking, and evidence-based reasoning in both textual and multimodal environments.
    For textual tasks, we use BrowseComp~\citep{wei2025browsecomp}, BrowseComp-ZH~\citep{zhou2025browsecompzhbenchmarkingwebbrowsing}, GAIA~\citep{mialon2023gaia}, Humanity's Last Exam~\citep{phan2025hle}, and xBench-DeepSearch~\citep{chen2025xbench}.
    For multimodal tasks, we use LiveVQA~\citep{fu2025livevqa}, MM-Search~\citep{wu2025mmsearch}, BrowseComp-VL~\citep{geng2025webwatcher}, RealXBench~\citep{hong2025deepeyesv2}, MM-BrowseComp~\citep{li2025mmbrowsecomp}, and HLE-VL~\citep{phan2025hle}.

    \item \textbf{Deep research report generation.}
    We evaluate the model's ability to collect information, synthesize evidence, and generate long-form research reports using DeepResearch Bench~\citep{du2025deepresearchbench}, DeepResearch Bench II~\citep{li2026deepresearchbench2}, and ResearchRubrics~\citep{sharma2025researchrubrics}.

    \item \textbf{Deep research instruction following.}
    We evaluate the model's ability to follow complex constraints during long-horizon research tasks using ComplexBench~\citep{wen2024complexbenchmarking} and our in-house DeepResearchIF benchmark.

    \item \textbf{File understanding and generation.}
    We evaluate file-centric tasks requiring attachment understanding, information extraction, tool-assisted processing, and artifact generation using the GAIA attachment subset~\citep{mialon2023gaia}, GTA~\citep{wang2024gta}, and our in-house FileSys benchmark.

    \item \textbf{Dynamic skill utilization.}
    We introduce the SkillsUse benchmark to evaluate whether models can understand external skill specifications, invoke appropriate skills, and complete complex tasks through skill-guided execution.

\end{itemize}

\textbf{In-House Benchmarks.}
While public benchmarks provide standardized evaluation protocols, they mainly focus on well-defined tasks with explicit objectives and cannot fully capture open-ended deep research scenarios.
Therefore, we further construct three in-house benchmarks to evaluate whether models can handle user-provided materials, satisfy complex requirements, interact with external capabilities, and produce verifiable deliverables.

\begin{itemize}
    \item \textbf{FileSys}
    contains 454 samples and evaluates whether models can generate usable deliverable files from natural-language requests.
    It covers DOCX, PDF, HTML, XLSX, SVG diagrams, data visualizations, and other structured or visual artifacts.
    FileSys reports two metrics: \textit{CodeExc}, which measures successful execution and file creation, and \textit{FileAns}, which evaluates whether the generated artifact semantically satisfies the task requirements.

    \item \textbf{DeepResearchIF}
    contains 900 examples across general, scientific, and industrial scenarios, and evaluates whether models can satisfy research-oriented constraints over task scope, source selection, evidence usage, analytical methods, assumptions, reasoning procedures, and long-form report generation.
    It reports strict sample-level accuracy and a constraint-level macro-average score over 9 top-level constraint categories and 26 fine-grained constraint types.

    \item \textbf{SkillsUse}
    contains 400 queries in No-attachment and Attachment settings, and evaluates whether models can discover relevant skills, read skill documentation, avoid distractors, follow skill-specific procedures, and complete tasks through skill-guided execution.
    Each trajectory is judged along Result, Execution, and Skill Usage dimensions with 12 fine-grained metrics.
\end{itemize}

Detailed descriptions of all benchmarks, including data composition, evaluation metrics, and protocols, are provided in Appendix~\ref{app:public_benchmarks} \& \ref{app:inhouse_benchmarks}.

\subsubsection{Baselines}

We compare S1-DeepResearch with a comprehensive set of baselines covering open-weight models, frontier proprietary models, and specialized deep research agents.

\begin{itemize}
    \item \textbf{Open-weight models.}
    We include open-weight models across different parameter scales, including Qwen3-32B, Qwen3-235B~\citep{yang2025qwen3}, Qwen3.5-397B~\citep{qwen2026qwen35}, GLM-5~\citep{glm2026glm5}, Kimi-K2.5~\citep{moonshot2026kimitwopointfive}, DeepSeek-V3.2~\citep{deepseek2026v32}, and MiniMax-M2.7~\citep{minimax2026m27}.

    \item \textbf{Closed-source frontier models.}
    We compare with leading proprietary models, including Doubao-Seed-2.0-Pro~\citep{seed2026seed20}, Gemini-3.1-Pro-Preview~\citep{google2026gemini31pro}, Claude-4.6-Sonnet-Thinking~\citep{anthropic2026claudesonnet46}, and GPT-5.2~\citep{openai2025gpt52}.

    \item \textbf{Specialized deep research agents.}
    We further compare with models and systems specifically optimized for deep research, including Gemini-DeepResearch~\citep{googledeepmind2026deepresearchmax}, OpenAI-DeepResearch~\citep{openai2025deepresearch}, UniScientist~\citep{unipat2026uniscientist}, Step-DeepResearch~\citep{hu2025stepdeepresearch}, REDSearcher~\citep{chu2026redsearcher}, Tongyi-DeepResearch~\citep{tongyi2025deepresearch}, MiroThinker-1.7 series~\citep{miromind2026mirothinker}, OpenSeeker-v1\citep{du2026openseeker}, OpenResearcher\citep{li2026openresearcher}, Vision-DeepResearch~\citep{huang2026visiondeepresearchincentivizingdeepresearchcapability}, Skywork-R1V4-30B~\citep{zhang2025skyworkr1v4agenticmultimodalintelligence}, and MM-DeepResearch~\citep{yao2026mmdeepresearchsimpleeffectivemultimodal}.
\end{itemize}

\subsubsection{Evaluation Settings}

To ensure fair comparison across different models and systems, we evaluate all baselines under a unified agent environment whenever applicable.
For tasks requiring external interactions, models are provided with the same set of atomic tools, including 
\textbf{web search},
\textbf{web visit},
\textbf{image search},
\textbf{academic search},
\textbf{file parsing},
\textbf{code execution},
\textbf{bash},
\textbf{image question answering},
and \textbf{video question answering}.
All models receive identical task inputs and follow the same evaluation protocols.

We use consistent inference settings across all experiments.
The temperature is set to 0.85, the top-p value is set to 0.95, and the repetition penalty is set to 1.1.
Each task allows up to 150 tool calls with a maximum context length of 128K tokens.

\begin{table}[t]
\centering
\caption{
Evaluation results on Textual long-horizon complex reasoning benchmarks.
Unless otherwise marked, scores are obtained from our evaluation under a unified tool configuration. $\dagger$ denotes officially reported scores, while $\ddagger$, $\S$, and $\P$ denote scores reported in MiroThinker\citep{miromind2026mirothinker}, GLM-5\citep{zai2026glm5blog}, and Nanbeige4.1-3B\citep{yang2026nanbeige4}, respectively.
\textbf{Overall} is the average score over available benchmark results, and the best result in each column is underlined.
}
\label{tab:long_horizon_complex_reasoning_text}
\scriptsize
\setlength{\tabcolsep}{4pt} 
\renewcommand{\arraystretch}{1.15} 

\begin{adjustbox}{max width=\linewidth}
\begin{tabular}{lcccccccc}
\toprule
\textbf{Model} & \textbf{Size} & \textbf{Train} & \textbf{GAIA(Text)} & \textbf{BrowseComp} & \textbf{BrowseComp-ZH} & \textbf{xBench-DeepSearch} & \textbf{HLE(Text)} & \textbf{Overall} \\
\midrule

\multicolumn{9}{c}{\textbf{\tiny Proprietary General Models}} \\
\midrule
Doubao-2.0-Pro & - & - & 78.6 & 77.3\officialscore & \underline{82.4}\officialscore & \underline{85.0} & 54.2\officialscore & \underline{75.5} \\
Gemini-3.1-Pro-Preview & - & - & 70.9 & \underline{85.9}\officialscore & 63.3 & 80.0 & 51.4\officialscore & 70.3 \\
Claude-4.6-Sonnet-Thinking & - & - & 71.8 & 74.7\officialscore & 49.4 & 76.0 & 49.0\officialscore & 64.2 \\
GPT-5.2 & - & - & 68.0 & 65.8\paperbscore & 76.1\paperbscore & 83.0 & 45.5\paperbscore & 67.7 \\

\midrule
\multicolumn{9}{c}{\textbf{\tiny Open-Source General Models}} \\
\midrule
Qwen3-32B & 32B & - & 30.2\papercscore & 3.2\papercscore & 7.3\papercscore & 39.0\papercscore & 9.3\papercscore & 17.8 \\
Qwen3-235B & 235B & - & 33.0 & 11.5 & 13.5 & 45.0 & 11.8\officialscore & 23.0 \\
Qwen3.5-397B & 397B & - & 73.8 & 78.6\officialscore & 70.3\officialscore & 84.0 & 48.3\officialscore & 71.0 \\
GLM-5 & 744B & - & 68.0 & 75.9\officialscore & 72.7\officialscore & 82.0 & 50.4\officialscore & 69.8 \\
Kimi-K2.5 & 1T & - & 68.9 & 74.9\officialscore & 62.3\officialscore & 78.0 & 52.2\officialscore & 67.3 \\
DeepSeek-V3.2 & 671B & - & 63.5\paperascore & 51.4\officialscore & 65.0\officialscore & 71.0\paperascore & 40.8\officialscore & 58.3 \\
MiniMax-M2.7 & 230B & - & \underline{81.6} & 76.1 & 56.4 & \underline{85.0} & 43.8 & 68.6 \\

\midrule
\multicolumn{9}{c}{\textbf{\tiny Specialized Deep Research Agentic Textual Models}} \\
\midrule
REDSearcher & 30B & Mid/SFT/RL & 80.1\officialscore & 42.1\officialscore & 49.8\officialscore & - & - & - \\
Tongyi-DeepResearch & 30B & Mid/SFT/RL & 70.9\officialscore & 43.4\officialscore & 46.7\officialscore & 75.0\officialscore & 32.9\officialscore & 53.8 \\
OpenSeeker-v1 & 30B & SFT & - & 29.5\officialscore & 48.4\officialscore & 74.0\officialscore & - & - \\
OpenResearcher & 30B & SFT & 64.1\officialscore & 26.3\officialscore & - & 65.0\officialscore & - & - \\
MiroThinker-1.7-mini & 30B & Mid/SFT/RL & - & 67.9\officialscore & 72.3\officialscore & - & 57.2\officialscore & - \\
MiroThinker-1.7 & 235B & Mid/SFT/RL & - & 74.0\officialscore & 75.3\officialscore & - & \underline{62.0}\officialscore & - \\

\midrule
\rowcolor{cyan!20}
S1-DeepResearch & 32B & SFT & 72.8 & 36.7 & 48.4 & 79.3 & 30.3 & 53.5 \\

\bottomrule
\end{tabular}
\end{adjustbox}

\end{table}

\subsection{Main Results}

As shown in Figure \ref{fig:overall_performance}, S1-DeepResearch, despite using only a 32B backbone, achieves strong performance across all five deep research dimensions after training on high-quality agentic workflows. Compared with the Qwen3-32B base model, it exhibits broad and consistent improvements, and remains competitive with substantially larger open-weight models as well as several closed-source frontier systems. These results suggest that agentic post-training enhances the model's general capabilities in planning, retrieval, reasoning, tool use, and end-to-end research workflow completion. Overall, S1-DeepResearch demonstrates that a 32B model can acquire strong end-to-end research-agent capabilities when trained on high-quality agentic trajectories.

\subsubsection{Long-Horizon Complex Reasoning}

\textbf{Textual.}
The results in Table \ref{tab:long_horizon_complex_reasoning_text} show that S1-DeepResearch substantially strengthens closed-ended long-horizon reasoning on textual tasks compared with Qwen3-32B, and even surpasses Qwen3-235B.
This indicates that the model can effectively handle tasks requiring persistent retrieval, query reformulation, evidence localization, and multi-hop verification.

S1-DeepResearch also reaches a competitive level against specialized deep research models and some general models, performing particularly well on real-world multi-step QA, Chinese multi-hop retrieval, and profession-oriented deep search tasks.
In particular, its strong performance on GAIA(Text) and xBench-DeepSearch further demonstrates that S1-DeepResearch has acquired solid textual long-horizon reasoning ability under realistic retrieval-intensive settings.
Meanwhile, the remaining gaps on more challenging benchmarks such as BrowseComp and HLE(Text) indicate the limitation of the current training recipe, suggesting that further reinforcement learning is needed to improve the model’s performance.

\subsubsection{Deep Research Report Generation}

\begin{table}[t]
\centering
\caption{
Evaluation results on long-form deep research report generation benchmarks.
Unless otherwise marked, scores are obtained from our evaluation under a unified tool configuration. $\dagger$ denotes officially reported scores, while $\ddagger$, $\S$, and $\P$ denote scores reported in DeepResearch Bench\citep{du2025deepresearchbench}, DeepResearch Bench II\citep{li2026deepresearchbench2}, and ResearchRubrics\citep{sharma2025researchrubrics}, respectively.
\textbf{Overall} is the average score over available benchmark results, and the best result in each column is underlined.
}
\label{tab:long_report_writing}
\scriptsize
\setlength{\tabcolsep}{4pt} 
\renewcommand{\arraystretch}{1.15} 

\begin{adjustbox}{max width=\linewidth}

\begin{tabular}{lcccccc}
\toprule
\textbf{Model} & \textbf{Size} & \textbf{Train} & \textbf{DeepResearchBench} & \textbf{DeepResearchBench II} & \textbf{ResearchRubrics} & \textbf{Overall} \\

\midrule
\multicolumn{7}{c}{\textbf{\tiny Proprietary Deep Research Agents}} \\
\midrule
Gemini-2.5-Pro-DeepResearch & - & - & 48.9\paperascore & 42.0\paperbscore & 61.5\papercscore & 51.1 \\
OpenAI-DeepResearch & - & - & 47.0\paperascore & 45.4\paperbscore & 59.7\papercscore & 50.5 \\

\midrule
\multicolumn{7}{c}{\textbf{\tiny Proprietary General Models}} \\
\midrule
Doubao-2.0-Pro & - & - & \underline{53.3}\officialscore & 39.6 & 50.7\officialscore & 47.9 \\
Gemini-3.1-Pro-Preview & - & - & 42.6 & 40.7 & 51.0 & 44.8 \\
Claude-4.6-Sonnet-Thinking & - & - & 46.7 & \underline{51.9} & 60.5 & \underline{53.0} \\
GPT-5.2 & - & - & 49.4 & 45.2 & 59.7 & 51.4 \\

\midrule
\multicolumn{7}{c}{\textbf{\tiny Open-Source General Models}} \\
\midrule
Qwen3-32B & 32B & - & 36.0 & 27.8 & 41.7 & 35.2 \\
Qwen3-235B & 235B & - & 38.4 & 29.7 & 43.9 & 37.3 \\
Qwen3.5-397B & 397B & - & 45.7 & 44.2 & 58.6 & 49.5 \\
GLM-5 & 744B & - & 46.9 & 44.5 & \underline{63.4} & 51.6 \\
Kimi-K2.5 & 1T & - & 45.5 & 44.8 & 60.9 & 50.4 \\
DeepSeek-V3.2 & 671B & - & 45.6 & 42.8 & 53.5 & 47.3 \\
MiniMax-M2.7 & 230B & - & 46.0 & 41.9 & 62.5 & 50.1 \\

\midrule
\multicolumn{7}{c}{\textbf{\tiny Specialized Deep Research Agentic Report-Writing Models}} \\
\midrule
Tongyi-DeepResearch & 30B & Mid/SFT/RL & - & 29.9\paperbscore & - & - \\
UniScientist & 30B & SFT & 46.0\officialscore & 48.0\officialscore & 59.9\officialscore & 51.3 \\
Step-DeepResearch & 32B & Mid/SFT/RL  & - & - & 61.4\officialscore & - \\

\midrule
\rowcolor{cyan!20}
S1-DeepResearch & 32B & SFT & 46.5 & 41.7 & 58.7 & 48.7 \\

\bottomrule
\end{tabular}
\end{adjustbox}

\end{table}

Table \ref{tab:long_report_writing} evaluates long-form deep research report generation, a setting that requires models to move beyond short-form question answering and produce structured, evidence-grounded, and coherent research-style outputs.

Compared with its backbone model, S1-DeepResearch shows clear improvements across report-writing benchmarks.
This indicates that the model can more effectively convert retrieved evidence into structured research-style outputs, suggesting improvements beyond factual retrieval alone.
Compared with larger open-source general models, S1-DeepResearch achieves competitive report-generation performance with a 32B backbone.
It outperforms Qwen3-235B in overall performance and attains results comparable to substantially larger models, including Qwen3.5-397B, Kimi-K2.5, and MiniMax-M2.7.
This comparison suggests that long-form research writing is not solely governed by model scale, but also depends on learning complete research workflows from evidence collection to report construction.
Among deep research agents, S1-DeepResearch substantially outperforms general-purpose agents such as Tongyi-DeepResearch, and remains competitive with report-specialized agents such as UniScientist and Step-DeepResearch.

\subsubsection{Deep Research Instruction Following}

\begin{table}[t]
\centering
\caption{
Evaluation results on deep research instruction-following benchmarks.
Unless otherwise marked, scores are obtained from our evaluation under a unified tool configuration.
\textbf{Overall} is the average score over available benchmark results, and the best result in each column is underlined.
}
\label{tab:instruction_following}
\scriptsize
\setlength{\tabcolsep}{4pt} 
\renewcommand{\arraystretch}{1.15} 


\begin{adjustbox}{max width=\linewidth}

\begin{tabular}{lcccccc}
\toprule
\multirow{2}{*}{\textbf{Model}} 
& \multirow{2}{*}{\textbf{Size}} 
& \multirow{2}{*}{\textbf{Train}} 
& \multicolumn{2}{c}{\textbf{DeepResearchIF}} 
& \multirow{2}{*}{\textbf{ComplexBench}} 
& \multirow{2}{*}{\textbf{Overall}} \\
\cmidrule(lr){4-5}
& & & \textbf{Query-Level Acc.} & \textbf{Constraint-Level Macro-Avg} & & \\
\midrule

\multicolumn{7}{c}{\textbf{\tiny Proprietary General Models}} \\
\midrule
Doubao-2.0-Pro & - & - & 23.4 & 77.7 & 87.4 & 55.4 \\
Gemini-3.1-Pro-Preview & - & - & 24.8 & 71.3 & \underline{87.6} & 56.2 \\
Claude-4.6-Sonnet-Thinking & - & - & 33.6 & 81.1 & 85.6 & 59.6 \\
GPT-5.2 & - & - & \underline{56.4} & \underline{89.2} & 86.7 & \underline{71.6} \\

\midrule
\multicolumn{7}{c}{\textbf{\tiny Open-Source General Models}} \\
\midrule
Qwen3-32B & 32B & - & 4.1 & 28.9 & 77.0 & 40.6 \\
Qwen3-235B & 235B & - & 7.4 & 54.9 & 79.8 & 43.6 \\
Qwen3.5-397B & 397B & - & 14.6 & 68.4 & 86.1 & 50.4 \\
GLM-5 & 744B & - & 10.8 & 62.7 & 85.0 & 47.9 \\
Kimi-K2.5 & 1T & - & 18.9 & 74.3 & 83.5 & 51.2 \\
DeepSeek-V3.2 & 671B & - & 25.8 & 76.0 & 83.9 & 54.8 \\
MiniMax-M2.7 & 230B & - & 11.0 & 63.9 & 79.2 & 45.1 \\

\midrule
\rowcolor{cyan!20}
S1-DeepResearch & 32B & SFT & 25.2 & 74.3 & 83.1 & 54.2 \\

\bottomrule
\end{tabular}
\end{adjustbox}

\end{table}

Table~\ref{tab:instruction_following} reports the results on deep research instruction-following benchmarks.
Compared with Qwen3-32B, S1-DeepResearch achieves substantial improvements across all metrics, increasing Query-Level Accuracy from 4.1 to 25.2 and Constraint-Level Macro-Average from 28.9 to 74.3.
S1-DeepResearch also achieves competitive performance among open-source general models despite using a 32B backbone, and outperforms several substantially larger models on Query-Level Accuracy.

These results indicate that S1-DeepResearch is better able to maintain research-oriented constraints throughout long-horizon agentic trajectories. The improvement is mainly driven by our instruction-oriented data construction, which injects nine categories of deep research constraints into agentic tasks and exposes the model to complex compositional requirements involving sources, reasoning, objectives, formats, execution, and context.

\subsubsection{File Understanding and Generation}

\begin{table}[t]
\centering
\caption{
Evaluation results on file understanding and generation benchmarks.
Unless otherwise marked, scores are obtained from our evaluation under a unified tool configuration. $\dagger$ denotes officially reported scores.
\textbf{Overall} is the average score over available benchmark results, and the best result in each column is underlined.
}
\label{tab:file_understanding_generation}
\scriptsize
\setlength{\tabcolsep}{4pt} 
\renewcommand{\arraystretch}{1.15} 

\begin{adjustbox}{max width=\linewidth}

\begin{tabular}{lcccccc}
\toprule
\textbf{Model} & \textbf{Size} & \textbf{Train} & \textbf{GAIA (File)} & \textbf{GTA} & \textbf{FileSys} & \textbf{Overall} \\
\midrule

\multicolumn{7}{c}{\textbf{\tiny Proprietary Deep Research Agents}} \\
\midrule
Doubao-2.0-Pro & - & - & 71.0 & 75.0 & \underline{77.5} & \underline{74.5} \\
Gemini-3.1-Pro-Preview & - & - & 77.4 & 75.0 & 39.2 & 63.9 \\
Claude-4.6-Sonnet-Thinking & - & - & \underline{79.0} & \underline{77.9} & 56.2 & 71.0 \\
GPT-5.2 & - & - & 62.9 & 73.8 & 48.7 & 61.8 \\

\midrule
\multicolumn{7}{c}{\textbf{\tiny Open-Source General Models}} \\
\midrule
Qwen3-32B & 32B & - & 24.2 & 70.3 & 44.7 & 46.4 \\
Qwen3-235B & 235B & - & 40.3 & 65.7 & 37.0 & 47.7 \\
Qwen3.5-397B & 397B & - & 67.7 & 73.3 & 53.3 & 64.8 \\
GLM-5 & 744B & - & 70.7 & 72.7 & 64.3 & 69.2 \\
Kimi-K2.5 & 1T & - & 67.7 & 71.5 & 42.7 & 60.7 \\
DeepSeek-V3.2 & 671B & - & 62.9 & 47.7 & 53.3 & 54.6 \\
MiniMax-M2.7 & 230B & - & 69.4 & 72.1 & 39.6 & 60.4 \\

\midrule
\rowcolor{cyan!20}
S1-DeepResearch & 32B & SFT & 62.9 & 68.0 & 69.3 & 66.7 \\

\bottomrule
\end{tabular}
\end{adjustbox}

\end{table}

Table \ref{tab:file_understanding_generation} reports the results on file understanding and generation benchmarks.
S1-DeepResearch consistently improves over its backbone across file-centric tasks, with the most notable gains on FileSys, where models must produce executable artifacts rather than only answer questions about uploaded files.
This suggests that file-oriented agentic training strengthens the model's ability to translate user requirements into concrete tool-supported operations and generate usable deliverables.
Moreover, S1-DeepResearch remains competitive with substantially larger general models, indicating that practical file understanding and generation benefit not only from model scale, but also from task-specific agentic trajectories.

\subsubsection{Dynamic Skill Utilization}

\begin{table}[t]
\centering
\caption{
Evaluation results on the SkillUse benchmark.
Unless otherwise marked, scores are obtained from our evaluation under a unified tool configuration.
\textbf{Overall} is the average score over the two attachment settings, and the best result in each column is underlined.
}
\label{tab:skill_using}
\scriptsize
\setlength{\tabcolsep}{4pt} 
\renewcommand{\arraystretch}{1.15} 


\begin{adjustbox}{max width=\linewidth}

\begin{tabular}{lcccc}
\toprule
\multirow{2}{*}{\textbf{Model}} 
& \multirow{2}{*}{\textbf{Size}} 
& \multicolumn{2}{c}{\textbf{SkillUse}} 
& \multirow{2}{*}{\textbf{Overall}} \\
\cmidrule(lr){3-4}
& & \textbf{w/ attachment} & \textbf{w/o attachment} & \\
\midrule

\multicolumn{5}{c}{\textbf{\tiny Proprietary Deep Research Agents}} \\
\midrule
Doubao-2.0-Pro & - & 70.3 & 76.9 & 73.6 \\
Gemini-3.1-Pro-Preview & - & 70.9 & \underline{78.1} & \underline{74.5} \\
Claude-4.6-Sonnet-Thinking & - & \underline{71.3} & 75.9 & 73.6 \\
GPT-5.2 & - & 55.6 & 69.7 & 62.7 \\

\midrule
\multicolumn{5}{c}{\textbf{\tiny Open-source General Models}} \\
\midrule
Qwen3-32B & 32B & 45.6 & 44.3 & 44.9 \\
Qwen3-235B & 235B & 46.0 & 46.2 & 46.1 \\
Qwen3.5-397B & 397B & 63.6 & 68.5 & 66.0 \\
GLM-5 & 744B & 61.9 & 71.9 & 66.9 \\
Kimi-K2.5 & 1T & 64.3 & 73.5 & 68.9 \\
DeepSeek-V3.2 & 671B & 70.1 & 71.6 & 70.8 \\
MiniMax-M2.7 & 230B & 60.3 & 67.2 & 63.7 \\

\midrule
\rowcolor{cyan!20}
S1-DeepResearch & 32B & 69.7 & 71.7 & 70.1 \\

\bottomrule
\end{tabular}
\end{adjustbox}

\end{table}

Table \ref{tab:skill_using} reports the results on the SkillUse benchmark, which evaluates whether models can solve tasks conditioned on dynamically provided skill specifications.
Compared with its backbone, S1-DeepResearch achieves consistent improvements in both attachment-free and attachment-based settings, indicating that the model is better able to ground task execution in external procedural knowledge rather than relying only on generic tool-use patterns.

S1-DeepResearch also reaches a comparable level to DeepSeek-V3.2 despite using a much smaller backbone.
This competitiveness is closely tied to our skill-oriented trajectory construction, which exposes the model to scenarios involving target and distractor skills, progressive skill specifications, and tool operations under skill-specific constraints.

\begin{figure*}[t]
    \centering
    \includegraphics[width=0.7\textwidth]{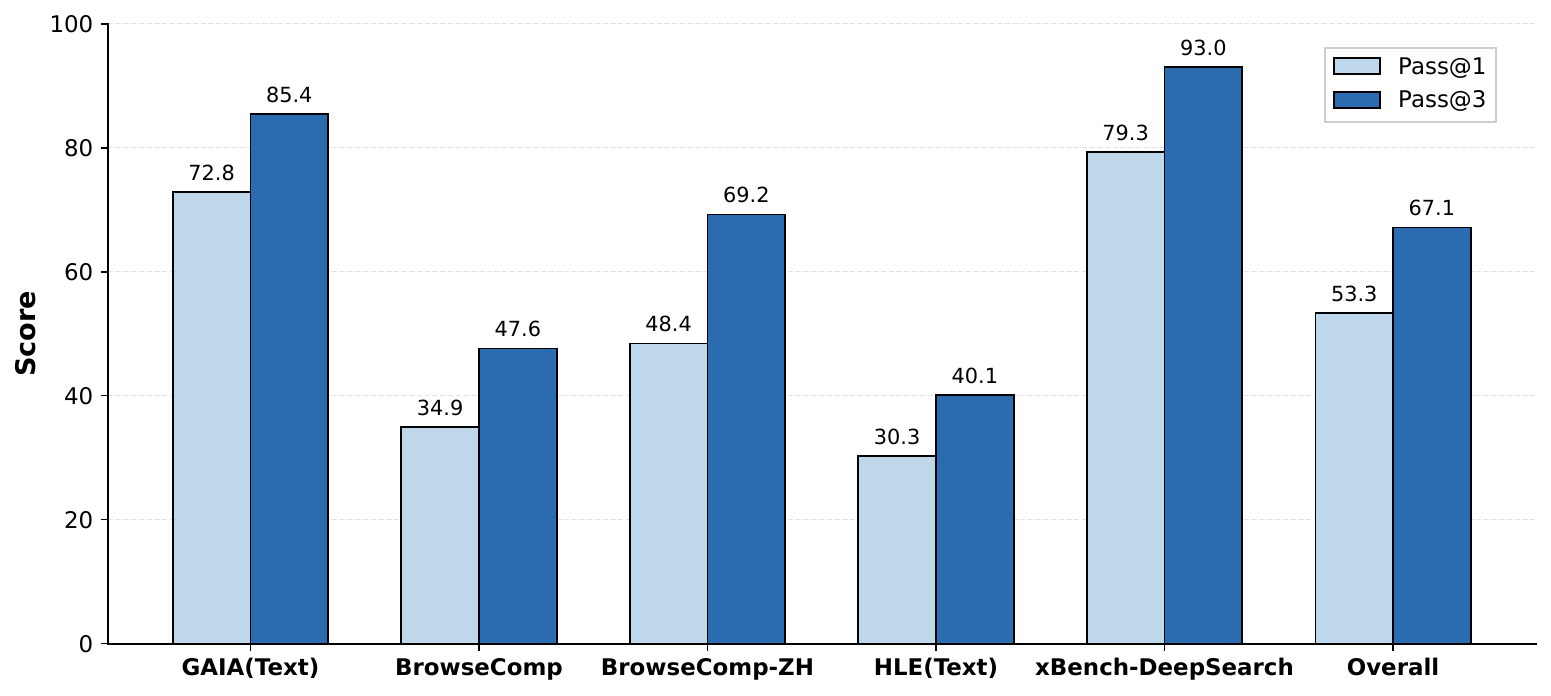}
    \caption{
    Test-time scaling performance of S1-DeepResearch on textual long-horizon complex reasoning benchmarks.
    We compare Pass@1 and Pass@3 across five benchmarks.
    }
    \label{fig:textual_test_time_scaling}
\end{figure*}

\begin{figure*}[t]
    \centering
    \includegraphics[width=0.8\textwidth]{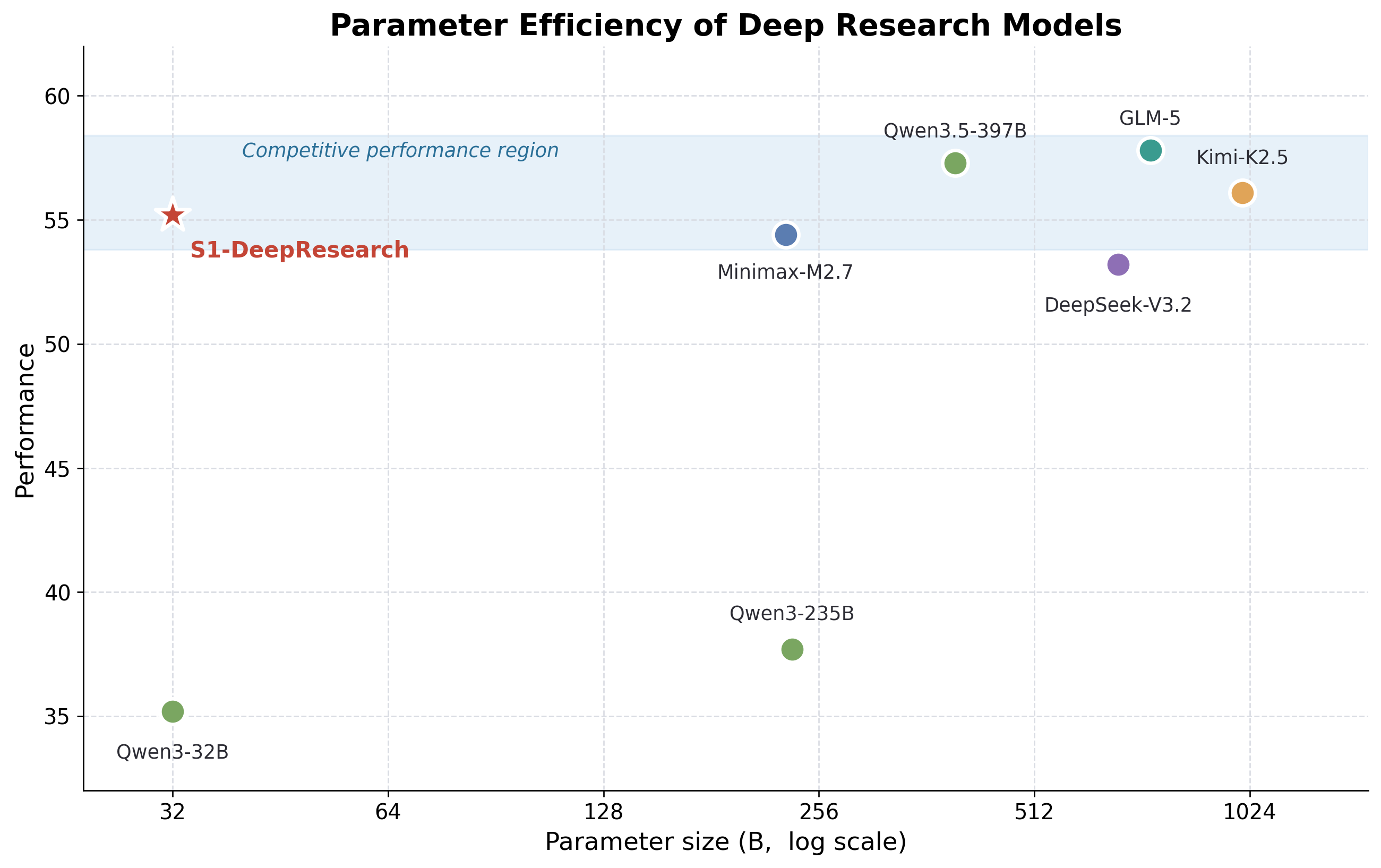}
    \caption{
    Parameter efficiency comparison of deep research models.
    S1-DeepResearch achieves competitive performance with a 32B backbone, 
    reaching a similar performance region to substantially larger models 
    while using roughly $30\times$ fewer parameters.
    }
    \label{fig:parameter_efficiency}
\end{figure*}

\subsection{Analysis}

\textbf{Test-Time Scaling.}
Figure \ref{fig:textual_test_time_scaling} shows that S1-DeepResearch exhibits strong test-time scaling potential on textual long-horizon reasoning tasks. The consistent improvement from Pass@1 to Pass@3 indicates that additional sampling rollouts help the model explore more diverse retrieval paths, recover from suboptimal intermediate decisions, and refine evidence coverage.

This behavior is particularly important for deep research tasks, where failures often arise from poor retrieval entry points, incomplete verification, or reasoning drift rather than a lack of basic capability. The results suggest that S1-DeepResearch does not merely perform well along a single deterministic trajectory, but can further benefit from increased test-time computation through richer exploration. This provides a solid foundation for tackling highly challenging closed-ended deep research tasks.

\textbf{Parameter Efficiency.}
Figure~\ref{fig:parameter_efficiency} analyzes the performance of S1-DeepResearch from the perspective of model scale.
Despite using only a 32B backbone, S1-DeepResearch reaches the competitive performance region of substantially larger models, including models with hundreds of billions to around one trillion parameters.
Compared with these large-scale baselines, it achieves comparable deep research performance with roughly $30\times$ fewer parameters.
These results indicate that strong deep research capability does not necessarily require proportional growth in model parameters.
Instead, the parameter efficiency of S1-DeepResearch is largely enabled by the proposed Agentic Data Construction System, which synthesizes high-quality trajectories across five core capability dimensions.
Rather than treating deep research as an extension of search-centric reasoning, our Agentic Data Construction System systematically transforms real-world research requirements into verifiable training trajectories across five core capability dimensions.
These trajectories jointly capture the essential structure of deep research workflows, including long-horizon exploration, evidence-grounded reasoning, report-level synthesis, constraint-aware instruction following, file-centric task completion, and dynamic skill utilization.

\subsection{Discussion on Multimodal Evaluation}

\begin{table}[t]
\centering
\caption{
Evaluation results on multimodal long-horizon complex reasoning benchmarks.
Unless otherwise marked, scores are obtained from our evaluation under a unified tool configuration. $\dagger$ denotes officially reported scores.
The best result in each column is underlined.
}
\label{tab:long_horizon_complex_reasoning_multimodal}
\scriptsize
\setlength{\tabcolsep}{4pt}
\renewcommand{\arraystretch}{1.15}

\begin{adjustbox}{max width=\linewidth}
\begin{tabular}{lcccccccc}
\toprule
\textbf{Model} & \textbf{Size} & \textbf{Train} & \textbf{LiveVQA} & \textbf{MM-Search} & \textbf{BrowseComp-VL} & \textbf{RealXBench} & \textbf{MM-BrowseComp} & \textbf{HLE-VL} \\
\midrule

\multicolumn{9}{c}{\textbf{\tiny Proprietary General Models}} \\
\midrule
Doubao-2.0-Pro & - & - & 50.9 & 50.9 & 48.4 & 32.5 & 27.2 & 17.5 \\
Gemini-3.1-Pro-Preview & - & - & 66.7 & 62.0 & 41.1 & 34.5 & 11.6 & \underline{25.1} \\
Claude-4.6-Sonnet-Thinking & - & - & \underline{81.3} & \underline{71.3} & \underline{55.6} & 30.4 & \underline{41.1} & 24.3 \\
GPT-5.2 & - & - & 56.7 & 56.7 & 45.6 & 30.4 & 22.3 & 20.8 \\

\midrule
\multicolumn{9}{c}{\textbf{\tiny Open-Source General Models}} \\
\midrule
Qwen3-235B & 235B & - & 52.0 & 38.6 & 32.8 & 32.6 & 6.7 & 9.6 \\
Qwen3-32B & 32B & - & 50.3 & 35.1 & 26.8 & 28.4 & 7.1 & 11.4 \\
Qwen3.5-397B & 397B & - & 75.7 & 55.6 & 45.4 & 32.0 & 27.2 & 18.7 \\
GLM-5 & 744B & - & 77.0 & 61.4 & 34.6 & 34.5 & 22.8 & 19.9 \\
Kimi-K2.5 & 1T & - & 56.3 & 57.9 & 37.1 & \underline{35.6} & 20.5 & 22.3 \\
DeepSeek-V3.2 & 671B & - & 50.7 & 46.8 & 30.3 & 23.7 & 31.7 & 14.9 \\
MiniMax-M2.7 & 230B & - & 69.7 & 56.1 & 40.1 & 32.0 & 20.1 & 13.1 \\

\midrule
\multicolumn{9}{c}{\textbf{\tiny Specialized Deep Research Agentic Multi-Modal Models}} \\
\midrule
Skywork-R1V4-30B & 30B & SFT & - & 66.1\officialscore & 38.4\officialscore & - & - & - \\
Vision-DeepResearch & 30B & SFT/RL & 77.6\officialscore & 69.6\officialscore & 53.7\officialscore & - & - & - \\
REDSearcher-MM-SFT & 30B & Mid/SFT & 78.5\officialscore & 70.3\officialscore & 55.3\officialscore & - & 25.3\officialscore & 24.2\officialscore \\
MM-DeepResearch & 32B & SFT/RL & 68.0\officialscore & 69.0\officialscore & 43.0\officialscore & - & - & - \\

\midrule
\rowcolor{cyan!20}
S1-DeepResearch & 32B & SFT & 67.7 & 54.4 & 39.1 & 31.4 & 19.2 & 15.2 \\

\bottomrule
\end{tabular}
\end{adjustbox}

\end{table}

\textbf{Tool-Augmented Multimodal Reasoning.}
Table \ref{tab:long_horizon_complex_reasoning_multimodal} reports the results on multimodal long-horizon reasoning benchmarks.
Unlike native multimodal large language models (MLLMs), S1-DeepResearch is built upon an LLM backbone and obtains multimodal capability through external image-understanding tools.
In this tool-augmented setting, visual inputs are first processed by perception tools and converted into textual observations, which are then consumed by the LLM for evidence integration, multi-step reasoning, and answer generation.

The results show that, by incorporating visual evidence from external perception tools together with search, browsing, and textual reasoning, S1-DeepResearch can address multimodal long-horizon tasks within the same agentic workflow.
Nevertheless, compared with native MLLMs and specialized multimodal research systems, S1-DeepResearch still exhibits clear performance gaps on several benchmarks.

These gaps are largely attributable to the architectural difference between tool-augmented perception and end-to-end multimodal representation learning.
Since visual information is exposed to S1-DeepResearch only through textual observations generated by external tools, fine-grained visual grounding, spatial reasoning, and cross-modal alignment remain indirectly accessible to the language model.
As a result, the multimodal results demonstrate both the potential and the limitations of extending an LLM-based research agent with external perception tools.

\section{Case Study}
\label{sec:case_study}

While the quantitative results demonstrate the overall effectiveness of S1-DeepResearch, they do not fully reveal how the model solves complex deep research tasks in practice. We therefore provide representative case studies to qualitatively examine the model's behavior across five capability dimensions: deep research report generation, dynamic skill utilization, file understanding and generation, deep research instruction following, and long-horizon complex reasoning. These cases are selected to cover different task forms, including professional analysis, scientific tool use, document generation, constrained information synthesis, text-based multi-hop reasoning, and multimodal open-world reasoning.

\textbf{Deep Research Report Generation.}
The report-generation case demonstrates that S1-DeepResearch can transform open-ended professional requests into structured, actionable, and domain-aware long-form reports. In the forensic accounting case for SMEs (Figure~\ref{fig:case_report_generation}), the model moves beyond generic financial advice and organizes the analysis into financial-statement diagnosis, early-warning indicators, cash-flow forecasting, intervention strategies, and post-intervention accounting controls. This indicates its ability to integrate domain knowledge from forensic accounting, financial risk management, and enterprise governance into a coherent analytical framework.

\textbf{Dynamic Skill Utilization.}
The skill-utilization case shows that S1-DeepResearch can solve tasks that require domain-specific tools or executable skills rather than plain text generation alone. In the TiO$_2$ surface-slab construction case (Figure~\ref{fig:case_dynamic_skill}), the model follows the workflow of materials modeling, including surface selection, slab generation, thickness and vacuum configuration, and structural verification. This reflects its ability to combine scientific knowledge with executable tool use and to refine intermediate results toward a valid technical output.

\textbf{File Understanding and Generation.}
The file-centric case illustrates that S1-DeepResearch can connect reasoning, document structuring, and artifact generation within a single workflow. In the geometry problem case (Figure~\ref{fig:case_file_generation}), the model first derives the circle radius from chord-length and distance constraints, then converts the solution into a structured HTML page with step-by-step explanations, labeled diagrams, and a highlighted final answer, and finally generates a PDF file. This demonstrates that the model can transform intermediate reasoning into polished user-facing documents and files.

\textbf{Deep Research Instruction Following.}
The instruction-following cases highlight the model's ability to satisfy complex, multi-part user requirements under explicit constraints. In the sustainable-architecture book-list case (Figure~\ref{fig:case_instruction_books}), the model must satisfy topical relevance, publication-year constraints, a fixed number of results, required metadata fields, and Markdown table formatting. In the IT-support training-plan case (Figure~\ref{fig:case_instruction_training}), it must produce a long-form professional document with mandatory sections, sufficient length, progressive curriculum design, hands-on exercises, assessments, and resource planning. Together, these examples show that S1-DeepResearch can maintain global instruction awareness over long outputs while preserving structure, completeness, and format consistency.

\textbf{Long-Horizon Complex Reasoning.}
The long-horizon reasoning cases demonstrate the model's ability to integrate dispersed clues across time, domains, and modalities. In the folic-acid discovery case (Figure~\ref{fig:case_reasoning_folic}), the model reconstructs a scientific timeline from pregnancy-related nutritional deficiency to yeast-extract factor, folic acid isolation, publication order, and crystallization year. In the railroad-and-bird-migration case (Figure~\ref{fig:case_reasoning_railroad}), it links ecological geography, migratory routes, railroad formation history, and later corporate acquisition events. In the multimodal geolocation case (Figure~\ref{fig:case_reasoning_geo}), it uses visual anchors to infer location, expands the analysis through external search, and recommends nearby places in a structured manner. These cases collectively indicate that S1-DeepResearch can perform sustained multi-hop reasoning over heterogeneous evidence.

\section{Limitations}
\label{sec:limitations}

Despite the strong performance achieved by S1-DeepResearch, several limitations remain.

\textbf{Limited Coverage of Coding-Oriented Tasks.}
S1-DeepResearch is primarily designed for real-world deep research scenarios, where agents are expected to perform long-horizon information seeking, evidence synthesis, instruction following, and deliverable generation.
As a result, the current framework places less emphasis on coding-oriented agentic capabilities, such as iterative program synthesis, execution-based debugging, repository-level code understanding, and software engineering task completion.
Compared with deep research workflows, these tasks place greater demands on code-environment interaction, executable feedback, program-level verification, and repository-scale context understanding.
Extending S1-DeepResearch toward coding-centric agentic scenarios remains an important direction for future work.

\textbf{Gap to Native Multimodal Reasoning.}
S1-DeepResearch currently supports multimodal tasks by using external visual understanding tools to convert visual inputs into textual observations, which are then integrated into the model’s reasoning process.
This tool-augmented design allows a text-centric research agent to handle multimodal evidence, but it still differs from native multimodal models that learn unified representations across visual and textual modalities.
Consequently, the current multimodal capability remains limited when tasks require fine-grained visual perception, spatial understanding, and direct cross-modal reasoning.
Such limitations are particularly evident in challenging open-world tasks such as visual geo-localization, where agents need to jointly interpret subtle visual cues, geographic context, and external evidence.
Extending the proposed data construction and training framework to native multimodal models is therefore an important direction for building stronger multimodal deep research agents.

\textbf{Exploration of Training Recipe.}
As shown in Table~\ref{tab:capability_comparison}, many existing deep research agents adopt more complex training recipes beyond supervised fine-tuning, including mid-training and reinforcement learning, to further enhance model capabilities.
Due to time and resource constraints, the current training procedure of S1-DeepResearch primarily relies on supervised fine-tuning over constructed agentic trajectories.
Although S1-DeepResearch is trained only with supervised fine-tuning, the proposed agentic data construction framework provides high-quality trajectories across multiple capability dimensions, enabling the model to achieve strong performance on a wide range of deep research benchmarks and approach leading proprietary frontier models on several challenging tasks.
Future work will investigate more advanced training strategies, such as agentic reinforcement learning and online preference optimization, to better exploit long-horizon feedback signals from interactive environments and further improve planning, decision making, and adaptive tool-use capabilities.

\section{Conclusion}

In this paper, we introduced S1-DeepResearch, a unified framework for deep research agents. To address the scarcity of high-quality deep research trajectories, we proposed a trajectory construction paradigm that combines closed-ended QA and open-ended exploration, together with a scalable data synthesis framework consisting of graph-grounded task formulation, agentic trajectory rollout, and multi-dimensional trajectory verification. Building upon this framework, we constructed a large-scale corpus of deep research trajectories covering long-chain complex reasoning, instruction following, report writing, file understanding and generation, and skills usage. Extensive experiments across a diverse set of deep research benchmarks demonstrate that S1-DeepResearch achieves strong performance on knowledge synthesis, complex reasoning, and planning-oriented tasks, establishing state-of-the-art results among open-source models of comparable scale and approaching proprietary frontier models on several challenging benchmarks.

\section{Contributions}

\textbf{Core Contributors:} Yao Dong$^{*}$, Xinglin Xiao, Liwei Dong, Xinlong Jin, Zhengbo Li, Heng Zhang, Duyun Wang, Nan Xu$^{*}$

\renewcommand{\thefootnote}{*}
\footnotetext{\sffamily Corresponding authors: Yao Dong and Nan Xu. Emails: yao.dong@wenge.com, nan.xu@wenge.com}

\clearpage

\bibliography{s1-deepresearch}
\bibliographystyle{colm2024_conference}

\newpage
\appendix
\section*{Appendix}
\section{Prompt Template}
\label{app:prompt_template}

\begin{tcolorbox}[title=System Prompt, breakable, colback=gray!5]
\small
\begin{verbatim}
You are a deep research assistant. Your core function is to conduct thorough, multi-source 
investigations into any topic. You must handle both broad, open-domain inquiries and queries within 
specialized academic fields. For every request, synthesize information from credible, diverse 
sources to deliver a comprehensive, accurate, and objective response. When you have gathered 
sufficient information and are ready to provide the definitive response, you must enclose the entire 
final answer within <answer></answer> tags.

# Tools

You may call one or more functions to assist with the user query.

You are provided with function signatures within <tools></tools> XML tags:
<tools>
{"type": "function", "function": {"name": "search", "description": "Perform Google web searches 
then returns a string of the top search results. Accepts multiple queries.", "parameters": {"type": 
"object", "properties": {"query": {"type": "array", "items": {"type": "string", "description": 
"The search query."}, "minItems": 1, "description": "The list of search queries."}}, "required": 
["query"]}}}
{"type": "function", "function": {"name": "visit", "description": "Visit webpage(s) and return the 
summary of the content.", "parameters": {"type": "object", "properties": {"url": {"type": "array", 
"items": {"type": "string"}, "description": "The URL(s) of the webpage(s) to visit. Can be a single 
URL or an array of URLs."}, "goal": {"type": "string", "description": "The specific information 
goal for visiting webpage(s)."}}, "required": ["url", "goal"]}}}
{"type": "function", "function": {"name": "PythonInterpreter", "description": "Executes Python 
code in a sandboxed environment, including writing or modifying files as needed. To use this tool, 
you must follow this format:\n1. The 'arguments' JSON object must be empty: {}.\n2. The Python code 
to be executed must be placed immediately after the JSON block, enclosed within <code> and </code> 
tags.\n\nIMPORTANT: Any output you want to see MUST be printed to standard output using the print() 
function.\n\nExample of a correct call:\n<tool_call>\n{"name": "PythonInterpreter", "arguments": 
{}}\n<code>\nimport numpy as np\n# Your code here\nprint(f"The result is: {np.mean([1,2,3])}")\n
</code>\n</tool_call>", "parameters": {"type": "object", "properties": {}, "required": []}}}
{"type": "function", "function": {"name": "google_scholar", "description": "Leverage Google 
Scholar to retrieve relevant information from academic publications. Accepts multiple queries. 
This tool will also return results from google search", "parameters": {"type": "object", 
"properties": {"query": {"type": "array", "items": {"type": "string", "description": "The search 
query."}, "minItems": 1, "description": "The list of search queries for Google Scholar."}}, 
"required": ["query"]}}}
{"type": "function", "function": {"name": "parse_file", "description": "This is a tool that can be 
used to parse multiple user uploaded local files or online files such as PDF, DOCX, PPTX, TXT, CSV, 
XLSX, DOC, ZIP, MP4, MP3.", "parameters": {"type": "object", "properties": {"files": {"type": 
"array", "items": {"type": "string"}, "description": "The online file's URLs or the user 
uploaded local file paths to be parsed."}}, "required": ["files"]}}}
{"type": "function", "function": {"name": "image_search", "description": "Search images by query 
and return a list of related images. Accepts multiple complementary search queries in a single 
call.", "parameters": {"type": "object", "properties": {"query": {"type": "array", "items": {
"type": "string", "description": "A single image search query string."}, "minItems": 1, 
"description": "Array of query strings. Multiple complementary search queries can be provided in 
one request for image search."}}, "required": ["query"]}}}
{"type": "function", "function": {"name": "ask_question_about_image", "description": "Identify 
image content and answer questions about one or more images.", "parameters": {"type": "object", 
"properties": {"image_path": {"type": "array", "items": {"type": "string", "description": "Local 
path or URL to an image file."}, "minItems": 1, "description": "Array of local paths or URLs to 
image files."}, "question": {"type": "string", "description": "Query about the image content."}}, 
"required": ["image_path", "question"]}}}
{"type": "function", "function": {"name": "ask_question_about_video", "description": "Ask a 
question about one or more videos.", "parameters": {"type":"object", "properties": {
"video_path": {"type": "array", "items": {"type": "string", "description": "Local path or URL to 
the video file."}, "minItems": 1, "description": "Array of local paths or URLs to video files."}, 
"question": {"type": "string", "description": "The question to ask about the video content."}}, 
"required": ["video_path", "question"]}}}
{"type": "function", "function": {"name": "bash", "description": "Execute ashell script in the 
current working directory. Use this tool to run one or more shell commands as a single script. Use 
relative paths by default. To use this tool, you must follow this format:\n1. The 'arguments' JSON 
object must be empty: {}.\n2. The Bash code to be executed must be placed immediatelyafter the JSON 
block, enclosed within <bash> and </bash> tags.\n\nExample of a correct call:\n<tool_call>\n{
"name": "bash", "arguments": {}}\n<bash>\n#Your bash code here\necho "Hello, world"\nls -l\n
</bash>\n</tool_call>", "parameters": {"type": "object", "properties": {}, "required": []}}}
</tools>

For each function call, return a json object with function name and arguments within <tool_call>
</tool_call> XML tags:
<tool_call>
{"name": <function-name>, "arguments": <args-json-object>}
</tool_call>

Current date: YYYY-MM-DD
\end{verbatim}
\end{tcolorbox}

\section{Details of Constraint Space}
\label{app:constraints_space}

To improve the controllability and complexity of generated research tasks, we design a nine-dimensional constraint space. The detailed definitions are as follows:

\begin{itemize}
    \item \textbf{Source Constraints.}
    Restrict the target scope of information retrieval and knowledge acquisition, including specific data sources, academic domains, or literature boundaries, ensuring domain consistency and high-quality evidence collection.

    \item \textbf{Argumentation Constraints.}
    Specify the required evidence organization, multi-perspective verification process, and completeness criteria of research conclusions to improve the rigor of generated tasks.

    \item \textbf{Reasoning Constraints.}
    Define the expected reasoning patterns when processing heterogeneous information, such as deductive reasoning, inductive summarization, causal analysis, and comparative reasoning.

    \item \textbf{Objective Constraints.}
    Define the core research objectives, key decision criteria, and evaluation requirements to mitigate objective drift during open-ended exploration.

    \item \textbf{Hypothetical Constraints.}
    Introduce predefined assumptions, physical boundaries, or logical conditions, restricting reasoning and analysis within a specified hypothesis space.

    \item \textbf{Output Format Constraints.}
    Specify the organization structure, formatting requirements, or markup specifications of final responses to ensure consistent structured outputs.

    \item \textbf{Output Scale Constraints.}
    Impose explicit quantitative boundaries on research coverage, analysis depth, and generation scale to control task complexity and granularity.

    \item \textbf{Execution Constraints.}
    Specify tool-use strategies, action sequence patterns, and resource budgets during exploration to improve the stability and controllability of agent execution.

    \item \textbf{Contextual Constraints.}
    Introduce temporal scopes, role perspectives, or application scenarios to enhance task realism and practical relevance.
\end{itemize}

\section{Tool Environment Details}
\label{app:tool_enviroment_details}

The tool environment in S1-DeepResearch consists of a set of tools that enable agents to interact with external resources, execute computations, and process multimodal information. Detailed tool descriptions and interface schemas are provided to specify the capabilities, input formats, and interaction protocols of each tool.

\subsection{Tool Descriptions}
\label{app:tool_descriptions}

The tool environment is designed as a collection of general-purpose atomic actions rather than task-specific research pipelines. Each tool provides a composable capability required in real-world research scenarios, enabling the agent to autonomously plan and combine different operations according to task requirements. These tools cover information acquisition, evidence extraction, file understanding, code execution, environment interaction, and multimodal analysis.

\textbf{Web Search} provides broad information discovery over the open web. It helps the model identify relevant sources, entities, terminology, timelines, viewpoints, and follow-up directions when the task scope is uncertain or open-ended.

\textbf{Web Visit} supports goal-directed reading of candidate web sources. It allows the model to inspect source content beyond snippets and extract evidence relevant to a specific research objective, such as factual claims, dates, provenance, experimental settings, or author positions.

\textbf{Academic Search} retrieves scholarly publications and research-oriented materials. It provides higher-quality evidence for literature review, method comparison, dataset tracing, benchmark analysis, and scientific claim verification than general web search alone.

\textbf{File Parsing} enables the model to process heterogeneous user-provided or online files, including PDFs, slides, spreadsheets, documents, archives, and transcripts. It converts these materials into model-consumable representations so that the agent can reason over private documents, uploaded files, and public sources jointly.

\textbf{Code Execution} supports deterministic computation, data analysis, and programmatic verification. It allows the model to clean tables, compute statistics, reproduce calculations, check data consistency, generate intermediate plots, and reduce numerical hallucinations in research tasks.

\textbf{Bash} provides shell-level access for file-system operations, command-line workflows, and executable research procedures. It enables the model to inspect directories, manipulate files, run scripts, invoke command-line tools, manage intermediate artifacts, and follow skill-specific execution instructions in a reproducible environment.

\textbf{Image Search} supports the discovery of visual evidence in open environments. It allows the model to retrieve images, charts, screenshots, maps, figures, or visual examples that complement text-based evidence.

\textbf{Image Question Answering} enables goal-directed interpretation of images. It allows the model to analyze visual materials with respect to the current research goal, such as reading charts, identifying objects, inspecting interfaces, or extracting evidence from figures and screenshots.

\textbf{Video Question Answering} enables goal-directed understanding of temporal or dynamic visual evidence. It allows the model to inspect videos and extract relevant actions, events, transitions, displayed information, or process-level evidence that cannot be captured by static images alone.

\subsection{Tool Schemas}
\label{app:tool_schemas}

\textbf{Tool Interface Compatibility.}
Due to differences in tool-calling protocols supported by different model serving backends, we provide two equivalent schema formats for code execution tools. Both formats share the same execution environment, permission settings, and runtime behavior, differing only in how tool inputs are serialized.
For models supporting native function-calling APIs, tool inputs are passed through standard JSON argument fields following the conventional function-call schema. For models evaluated with customized inference frameworks, we additionally adopt a XML-based interface, where the executable content is placed after the tool-call object and enclosed within dedicated tags (e.g., \texttt{<bash>} or \texttt{<code>}). This adaptation ensures compatibility across different serving backends while preserving identical tool functionality.


\begin{tcolorbox}[
  title=Web Search,
  colback=cyan!2,
  colframe=cyan!35!black,
  colbacktitle=cyan!12,
  coltitle=black,
  fonttitle=\bfseries,
  boxrule=0.45pt,
  arc=1.2pt,
  left=4pt,
  right=4pt,
  top=4pt,
  bottom=4pt
]
\begin{lstlisting}
{
    "type": "function",
    "function": {
        "name": "search",
        "description": "Perform Google web searches then returns a string of the top search results. Accepts multiple queries.",
        "parameters": {
            "type": "object",
            "properties": {
                "query": {
                    "type": "array",
                    "items": {
                        "type": "string",
                        "description": "The search query."
                    },
                    "minItems": 1,
                    "description": "The list of search queries."
                }
            },
            "required": ["query"]
        }
    }
}
\end{lstlisting}
\end{tcolorbox}

\begin{tcolorbox}[
  title=Web Visit,
  colback=cyan!2,
  colframe=cyan!35!black,
  colbacktitle=cyan!12,
  coltitle=black,
  fonttitle=\bfseries,
  boxrule=0.45pt,
  arc=1.2pt,
  left=4pt,
  right=4pt,
  top=4pt,
  bottom=4pt
]
\begin{lstlisting}
{
    "type": "function",
    "function": {
        "name": "visit",
        "description": "Visit webpage(s) and return the summary of the content.",
        "parameters": {
            "type": "object",
            "properties": {
                "url": {
                    "type": "array",
                    "items": {
                        "type": "string"
                    },
                    "description": "The URL(s) of the webpage(s) to visit. Can be a single URL or an array of URLs."
                },
                "goal": {
                    "type": "string",
                    "description": "The specific information goal for visiting webpage(s)."
                }
            },
            "required": ["url", "goal"]
        }
    }
}
\end{lstlisting}
\end{tcolorbox}

\begin{tcolorbox}[
  title=Academic Search,
  colback=cyan!2,
  colframe=cyan!35!black,
  colbacktitle=cyan!12,
  coltitle=black,
  fonttitle=\bfseries,
  boxrule=0.45pt,
  arc=1.2pt,
  left=4pt,
  right=4pt,
  top=4pt,
  bottom=4pt
]
\begin{lstlisting}
{
    "type": "function",
    "function": {
        "name": "google_scholar",
        "description": "Leverage Google Scholar to retrieve relevant information from academic publications. Accepts multiple queries. This tool will also return results from google search.",
        "parameters": {
            "type": "object",
            "properties": {
                "query": {
                    "type": "array",
                    "items": {
                        "type": "string",
                        "description": "The search query."
                    },
                    "minItems": 1,
                    "description": "The list of search queries for Google Scholar."
                }
            },
            "required": ["query"]
        }
    }
}
\end{lstlisting}
\end{tcolorbox}

\begin{tcolorbox}[
  title=File Parsing,
  colback=cyan!2,
  colframe=cyan!35!black,
  colbacktitle=cyan!12,
  coltitle=black,
  fonttitle=\bfseries,
  boxrule=0.45pt,
  arc=1.2pt,
  left=4pt,
  right=4pt,
  top=4pt,
  bottom=4pt
]
\begin{lstlisting}
{
    "type": "function",
    "function": {
        "name": "parse_file",
        "description": "This is a tool that can be used to parse multiple user uploaded local files or online files such as PDF, DOCX, PPTX, TXT, CSV, XLSX, DOC, ZIP, MP4, MP3.",
        "parameters": {
            "type": "object",
            "properties": {
                "files": {
                    "type": "array",
                    "items": {
                        "type": "string"
                    },
                    "description": "The online file's URLs or the user uploaded local file paths to be parsed."
                }
            },
            "required": ["files"]
        }
    }
}
\end{lstlisting}
\end{tcolorbox}

\begin{tcolorbox}[
  title=Code Execution (Standard Function Calling),
  colback=cyan!2,
  colframe=cyan!35!black,
  colbacktitle=cyan!12,
  coltitle=black,
  fonttitle=\bfseries,
  boxrule=0.45pt,
  arc=1.2pt,
  left=4pt,
  right=4pt,
  top=4pt,
  bottom=4pt
]
\begin{lstlisting}
{
    "type": "function",
    "function": {
        "name": "execute_code",
        "description": "Execute a given code snippet, such as processing data code, training a machine learning or deep learning model code, analysising data code, executing a workflow, etc.",
        "strict": true,
        "parameters": {
            "properties": {
                "code": {
                    "type": "string",
                    "description": "The input code to the Code Interpreter tool call."
                }
            },
            "required": ["code"],
            "type": "object",
            "additionalProperties": false
        }
    }
}
\end{lstlisting}
\end{tcolorbox}

\begin{tcolorbox}[
  title=Code Execution (XML-based Execution),
  colback=cyan!2,
  colframe=cyan!35!black,
  colbacktitle=cyan!12,
  coltitle=black,
  fonttitle=\bfseries,
  boxrule=0.45pt,
  arc=1.2pt,
  left=4pt,
  right=4pt,
  top=4pt,
  bottom=4pt
]
\begin{lstlisting}
{
    "type": "function",
    "function": {
        "name": "PythonInterpreter",
        "description": "Executes Python code in a sandboxed environment, including writing or modifying files as needed. To use this tool, you must follow this format:\n1. The 'arguments' JSON object must be empty: {}.\n2. The Python code to be executed must be placed immediately after the JSON block, enclosed within <code> and </code> tags.\n\nIMPORTANT: Any output you want to see MUST be printed to standard output using the print() function.\n\nExample of a correct call:\n<tool_call>\n{\"name\": \"PythonInterpreter\", \"arguments\": {}}\n<code>\nimport numpy as np\n# Your code here\nprint(f\"The result is: {np.mean([1,2,3])}\")\n</code>\n</tool_call>",
        "parameters": {
            "type": "object",
            "properties": {},
            "required": []
        }
    }
}
\end{lstlisting}
\end{tcolorbox}

\begin{tcolorbox}[
  title=Bash (Standard Function Calling),
  colback=cyan!2,
  colframe=cyan!35!black,
  colbacktitle=cyan!12,
  coltitle=black,
  fonttitle=\bfseries,
  boxrule=0.45pt,
  arc=1.2pt,
  left=4pt,
  right=4pt,
  top=4pt,
  bottom=4pt
]
\begin{lstlisting}
{
    "type": "function",
    "function": {
        "name": "bash",
        "description": "Execute a shell script in the current working directory. Use this tool to run one or more shell commands as a single script. Use relative paths by default. The Bash code to be executed must be provided in the `command` parameter.",
        "parameters": {
            "type": "object",
            "properties": {
                "command": {
                    "type": "string",
                    "description": "The Bash script or shell commands to execute in the current working directory."
                }
            },
            "required": ["command"]
        }
    }
}
\end{lstlisting}
\end{tcolorbox}

\begin{tcolorbox}[
  title=Bash (XML-based Execution),
  colback=cyan!2,
  colframe=cyan!35!black,
  colbacktitle=cyan!12,
  coltitle=black,
  fonttitle=\bfseries,
  boxrule=0.45pt,
  arc=1.2pt,
  left=4pt,
  right=4pt,
  top=4pt,
  bottom=4pt
]
\begin{lstlisting}
{
    "type": "function",
    "function": {
        "name": "bash",
        "description": "Execute a shell script in the current working directory. Use this tool to run one or more shell commands as a single script. Use relative paths by default. To use this tool, you must follow this format:\n1. The 'arguments' JSON object must be empty: {}.\n2. The Bash code to be executed must be placed immediatelyafter the JSON block, enclosed within <bash> and </bash> tags.\n\nExample of a correct call:\n<tool_call>\n{\"name\": \"bash\", \"arguments\": {}}\n<bash>\n#Your bash code here\necho \"Hello, world\"\nls -l\n</bash>\n</tool_call>",
        "parameters": {
            "type": "object",
            "properties": {},
            "required": []
        }
    }
}
\end{lstlisting}
\end{tcolorbox}

\begin{tcolorbox}[
  title=Image Search,
  colback=cyan!2,
  colframe=cyan!35!black,
  colbacktitle=cyan!12,
  coltitle=black,
  fonttitle=\bfseries,
  boxrule=0.45pt,
  arc=1.2pt,
  left=4pt,
  right=4pt,
  top=4pt,
  bottom=4pt
]
\begin{lstlisting}
{
    "type": "function",
    "function": {
        "name": "image_search",
        "description": "Search images by query and return a list of related images. Accepts multiple complementary search queries in a single call.",
        "parameters": {
            "type": "object",
            "properties": {
                "query": {
                    "type": "array",
                    "items": {
                        "type": "string",
                        "description": "A single image search query string."
                    },
                    "minItems": 1,
                    "description": "Array of query strings. Multiple complementary search queries can be provided in one request for image search."
                }
            },
            "required": ["query"]
        }
    }
}
\end{lstlisting}
\end{tcolorbox}

\begin{tcolorbox}[
  title=Image Question Answering,
  colback=cyan!2,
  colframe=cyan!35!black,
  colbacktitle=cyan!12,
  coltitle=black,
  fonttitle=\bfseries,
  boxrule=0.45pt,
  arc=1.2pt,
  left=4pt,
  right=4pt,
  top=4pt,
  bottom=4pt
]
\begin{lstlisting}
{
    "type": "function",
    "function": {
        "name": "ask_question_about_image",
        "description": "Identify image content and answer questions about one or more images.",
        "parameters": {
            "type": "object",
            "properties": {
                "image_path": {
                    "type": "array",
                    "items": {
                        "type": "string",
                        "description": "Local path or URL to an image file."
                    },
                    "minItems": 1,
                    "description": "Array of local paths or URLs to image files."
                },
                "question": {
                    "type": "string",
                    "description": "Query about the image content."
                }
            },
            "required": ["image_path", "question"]
        }
    }
}
\end{lstlisting}
\end{tcolorbox}

\begin{tcolorbox}[
  title=Video Question Answering,
  colback=cyan!2,
  colframe=cyan!35!black,
  colbacktitle=cyan!12,
  coltitle=black,
  fonttitle=\bfseries,
  boxrule=0.45pt,
  arc=1.2pt,
  left=4pt,
  right=4pt,
  top=4pt,
  bottom=4pt
]
\begin{lstlisting}
{
    "type": "function",
    "function": {
        "name": "ask_question_about_video",
        "description": "Ask a question about one or more videos.",
        "parameters": {
            "type": "object",
            "properties": {
                "video_path": {
                    "type": "array",
                    "items": {
                        "type": "string",
                        "description": "Local path or URL to the video file."
                    },
                    "minItems": 1,
                    "description": "Array of local paths or URLs to video files."
                },
                "question": {
                    "type": "string",
                    "description": "The question to ask about the video content."
                }
            },
            "required": ["video_path", "question"]
        }
    }
}
\end{lstlisting}
\end{tcolorbox}

\section{Public Benchmarks}
\label{app:public_benchmarks}

\textbf{Long-Horizon Complex Reasoning.} We evaluate the model on complex multi-hop deep research tasks with verifiable answers. To comprehensively assess deep research ability across different input modalities, we conduct evaluations under both text-only and multimodal input settings.

Under the text-only setting, we use the following benchmarks:

\begin{itemize}
    \item \textbf{BrowseComp}~\citep{wei2025browsecomp}, which evaluates long-horizon web browsing and information-seeking ability in English. It contains 1,266 questions that require persistent retrieval of hidden and interwoven information from the Internet, with short answers that can be automatically compared against references.
\end{itemize}

\begin{itemize}
    \item \textbf{BrowseComp-ZH}~\citep{zhou2025browsecompzhbenchmarkingwebbrowsing}, which extends this setting to the Chinese Internet. It contains 289 challenging multi-hop questions across 11 domains. Each question is reverse-constructed from an objective and verifiable short answer.
\end{itemize}

\begin{itemize}
    \item \textbf{GAIA (Text-Only)}~\citep{mialon2023gaia}, a general AI assistant benchmark containing real-world problems that require reasoning, web browsing, and tool use. In our evaluation, we use only its text-only subset.
\end{itemize}

\begin{itemize}
    \item \textbf{Humanity’s Last Exam (HLE, Text-Only)}~\citep{phan2025hle}, evaluated under the text-only setting, which measures the model’s ability to reason over and answer frontier closed-ended knowledge questions across mathematics, humanities, natural sciences, and other disciplines.
\end{itemize}

\begin{itemize}
    \item \textbf{xBench-DeepSearch}~\citep{chen2025xbench}, which evaluates information retrieval, task planning, and deep search ability in profession-aligned real-world settings.
\end{itemize}

Under the multimodal input setting, we further use the following benchmarks:

\begin{itemize}
    \item \textbf{LiveVQA}~\citep{fu2025livevqa}
    , a 300-example test subset sampled from LiveVQA following related evaluation settings. It evaluates the model’s ability to understand, retrieve, and answer questions about real-time visual knowledge.
\end{itemize}

\begin{itemize}
    \item \textbf{MM-Search}~\citep{jiang2024mmsearchbenchmarkingpotentiallarge}, which evaluates large models as multimodal search engines for handling image-text queries and web information. The benchmark covers key stages in multimodal search, including query understanding, search rewriting, result filtering, information aggregation, and answer generation.
\end{itemize}

\begin{itemize}
    \item \textbf{BrowseComp-VL}~\citep{geng2025webwatcher}, derived from WebWatcher-related research, which focuses on complex browsing-style question answering tasks involving both visual and textual information.
\end{itemize}

\begin{itemize}
    \item \textbf{RealXBench}~\citep{hong2025deepeyesv2}, which targets real-world multimodal reasoning tasks and covers visual perception, external search, information integration, and complex reasoning.
\end{itemize}

\begin{itemize}
    \item \textbf{MM-BrowseComp}~\citep{li2025mmbrowsecomp}, which evaluates browsing agents in mixed web environments containing images, videos, and text. It focuses on complex information-seeking tasks in realistic browsing scenarios.
\end{itemize}

\begin{itemize}
    \item \textbf{HLE-VL}~\citep{phan2025hle}, a subset of Humanity’s Last Exam containing image-based examples. It tests more challenging multidisciplinary visual knowledge question answering.
\end{itemize}

\textbf{Deep Research \& Long-form Report Writing.} We evaluate on DeepResearch Bench~\citep{du2025deepresearchbench}), ResearchRubrics~\citep{sharma2025researchrubrics}, and DeepResearch Bench II~\citep{li2026deepresearchbench2}. DeepResearch Bench includes 100 PhD-level tasks across 22 fields and introduces evaluation protocols for report quality as well as citation effectiveness and accuracy. ResearchRubrics provides realistic domain-diverse prompts with over 2,500 expert-written rubrics to measure factual grounding, reasoning soundness, and clarity. DeepResearch Bench II contains 132 grounded research tasks spanning 22 domains, evaluated using 9,430 fine-grained binary rubrics derived from expert-written investigative reports, covering information recall, analysis, and presentation.

\textbf{Deep Research Instruction Following.} We evaluate on ComplexBench~\citep{wen2024complexbenchmarking}. It focuses on multi-constraint composition in complex instructions, covering 4 constraint types, 19 constraint dimensions, and 4 composition types. It combines rule-based and LLM-based automatic evaluation to test whether models can simultaneously satisfy multiple formatting, content, logical, and semantic constraints.

\textbf{File Understanding and Generation.} We adopt the GAIA~\citep{mialon2023gaia} attachment subset and GTA~\citep{wang2024gta} as benchmarks. From the full GAIA  dataset, we select 62 samples with attachment inputs to build GAIA (File). From the full GTA dataset, we extract 172 tasks that are independent of specific tools and can be completed through general-purpose tools, forming GTA, which is used to evaluate the general tool-based problem-solving capability for attachment input tasks.

\section{In-House Benchmarks}
\label{app:inhouse_benchmarks}

\subsection{FileSys}
\label{app:filesys}

\textbf{Evaluation Goal.}
FileSys is an in-house benchmark designed to evaluate artifact generation in deep research scenarios.
In many realistic research workflows, the expected output is not limited to a plain-text answer, but may instead take the form of a deliverable artifact, such as a DOCX report, PDF document, HTML page, XLSX spreadsheet, figure, vector graphic, or other attachment-style output.
For example, users may ask an agent to produce a research report, construct a data spreadsheet, generate a visualization, design a webpage prototype, create a vector diagram, or revise a previously generated PDF through follow-up interactions.
FileSys therefore evaluates whether a model can transform a natural-language request into an executable artifact-generation process and ultimately produce a usable file with task-relevant content.

The benchmark assesses artifact generation from two complementary perspectives: execution behavior and content correctness.
The former examines whether the model actually triggers and completes the file-generation process, while the latter evaluates whether the generated artifact satisfies the semantic requirements of the task.

\textbf{Data Composition.}
FileSys contains 454 test examples covering mainstream artifact formats, including DOCX, PDF, HTML, SVG, and XLSX.
The benchmark is designed to evaluate artifact generation across several representative file-output scenarios:

\begin{itemize}
    \item \textbf{Document-style deliverables.}
    DOCX generation accounts for approximately 27\% of the benchmark and evaluates the model's ability to produce formal documents, research reports, manuals, and multi-section textual artifacts.
    PDF generation is evaluated in both single-turn and multi-turn settings.
    Direct PDF generation accounts for approximately 18\%, while two-turn PDF generation accounts for another 17\%.
    The multi-turn setting further examines whether the model can maintain contextual consistency and perform controllable revisions after follow-up user instructions.

    \item \textbf{Web and page-level outputs.}
    HTML generation accounts for approximately 17\% of the benchmark.
    These examples evaluate whether the model can organize structured content into a coherent page-level artifact, including information hierarchy, layout structure, and presentation-oriented content arrangement.

    \item \textbf{Graphical and structured artifacts.}
    FileSys also includes tasks beyond conventional document generation.
    Vector icons account for approximately 7\%, simple vector animations for approximately 5\%, chemical molecule vector diagrams for approximately 3\%, and XLSX spreadsheets for approximately 3\%.
    These examples test the model's ability to produce visual representations, encode graphical structures, and generate structured tabular files.
\end{itemize}

Overall, FileSys covers a broad range of artifact-generation scenarios, including textual document generation, multi-turn revision, page construction, visual expression, and structured data output.

\textbf{Evaluation Metrics.}
FileSys adopts a hierarchical evaluation protocol with two core metrics: \textit{CodeExc} and \textit{FileAns}.

\textit{CodeExc} measures whether the model successfully completes the artifact-generation process.
Specifically, it checks whether the model invokes the required code execution or file-generation operations, whether the execution terminates normally, whether the target file is created, and whether the generated file type matches the task requirement.
This metric reflects the model's end-to-end execution capability from instruction understanding to tool invocation and result materialization.
It serves as a prerequisite for subsequent content evaluation: if a model fails to trigger file generation, encounters an execution error, terminates abnormally, or does not produce the expected file type, the corresponding example is not further evaluated by \textit{FileAns}.

\textit{FileAns} measures the semantic correctness of the generated artifact after successful file generation.
Given a generated file, the evaluator compares the generated code, execution result, and artifact content against the standard code and reference artifact in the reference trajectory.
An LLM-as-a-Judge is then used to assess whether the core semantic content satisfies the task requirements.
Unlike visual-quality-oriented evaluation, \textit{FileAns} does not emphasize presentation details such as style, layout, CSS design, or visual aesthetics.
Instead, it focuses on content-level consistency, including whether key answers, entities, events, objects, logical relations, data conclusions, and task-specific requirements are correctly expressed.
Thus, \textit{FileAns} measures the reliability and usefulness of the generated artifact at the semantic level.

\textbf{Evaluation Protocol.}
FileSys adopts a two-stage evaluation protocol that first verifies artifact generation and then assesses content correctness.
In the first stage, \textit{CodeExc} is evaluated based on execution logs and file-system checks.
The evaluator verifies whether the execution process terminates successfully, whether the target file is generated, and whether the generated file format is consistent with the user request.
Examples that fail \textit{CodeExc} are excluded from subsequent content-level evaluation.

In the second stage, \textit{FileAns} is evaluated for examples that successfully generate the required artifact.
The evaluator parses the generated artifact, the corresponding generation code, and the reference trajectory, and then uses an LLM-based judge to assess semantic alignment with the reference artifact.
The assessment focuses on whether the generated artifact preserves the key information and satisfies the task requirements, rather than on superficial presentation details.
We report \textit{FileAns} as the primary metric and \textit{CodeExc} as the artifact-generation success rate.

This protocol enables FileSys to distinguish execution failures from content-level failures.
Such a distinction is essential for evaluating deep research agents, as artifact-generation tasks require both successful tool-based materialization and semantically correct deliverables.

\subsection{DeepResearchIF}
\label{app:deepresearchif}

\textbf{Evaluation Goal.}
DeepResearchIF is an in-house benchmark for evaluating instruction following in deep research scenarios.
Different from general instruction-following benchmarks that primarily emphasize surface-level requirements, such as length, format, keywords, or simple constraint combinations, DeepResearchIF focuses on research-oriented constraints in long-horizon and evidence-intensive tasks.
These constraints may involve task scope, source selection, evidence usage, analytical framework, reasoning procedure, assumptions, output organization, and report-generation requirements.

DeepResearchIF evaluates whether a model can accurately interpret and satisfy compositional user instructions during open-ended research.
The benchmark is designed to provide a diagnostic evaluation setting in which failures caused by inadequate information acquisition or analysis can be distinguished from failures caused by constraint violations.

\textbf{Data Composition.}
DeepResearchIF contains 900 test examples across three representative scenarios: general research, scientific research, and industrial research, with 300 examples in each scenario.
Each example is derived from a real research instruction and contains one or more explicit constraints.
To characterize instruction-following requirements in deep research tasks, we define a constraint taxonomy consisting of 9 top-level categories and 26 fine-grained types:

\begin{itemize}
    \item \textbf{Information-scope constraints} specify sources, time ranges, geographic regions, or data types.
    \item \textbf{Evidence-support constraints} require conclusions to be supported by verifiable evidence.
    \item \textbf{Reasoning-method constraints} specify analytical frameworks, methods, or logical paradigms.
    \item \textbf{Goal-orientation constraints} define judgment criteria, decision objectives, or optimization preferences.
    \item \textbf{Assumption constraints} require explicit treatment of uncertainty, scenarios, or external conditions.
    \item \textbf{Output-format constraints} specify organization, structure, expression format, or required components.
    \item \textbf{Output-scale constraints} control length, granularity, number of sections, or coverage depth.
    \item \textbf{Execution-mechanism constraints} specify tool usage, search strategies, procedural requirements, or automated operations.
    \item \textbf{Task-context constraints} define the audience, role perspective, business environment, or application scenario.
\end{itemize}

The constraint taxonomy covers both presentation-level requirements and task-level research requirements, including task boundary definition, evidence grounding, analytical procedure, execution mechanism, and contextual adaptation.
It therefore supports a more fine-grained evaluation of instruction following beyond format compliance.

\textbf{Evaluation Metrics.}
DeepResearchIF uses strict sample-level accuracy as the primary metric.
An example is marked correct only when all associated constraints are satisfied; otherwise, it is marked incorrect.
The final score is computed as the proportion of correctly completed examples.

Strict sample-level accuracy reflects the compositional nature of instruction following in deep research tasks.
A response that violates a key evidence boundary, analytical framework, structural requirement, or task context may fail to satisfy the user request even if other parts of the task are completed.
In addition to the primary metric, constraint-level annotations enable category-level diagnostic analysis across different types of research-oriented constraints.

\textbf{Evaluation Protocol.}
For each test example, the user instruction is decomposed into independently judgeable constraint items according to the predefined taxonomy.
Each constraint is annotated with its category, judgment criterion, and satisfaction condition.
Model outputs are evaluated at the constraint level and then aggregated into the final sample-level result.

Rule-based checks are applied to constraints that can be deterministically verified, such as output format, word count, number of sections, or number of citations.
LLM-based judges are used for semantic constraints, including analytical-framework usage, goal orientation, evidence support, and contextual adaptation.
Constraint-level judgments are retained for fine-grained error analysis across constraint categories and types.

\subsection{SkillsUse}
\label{app:skillsuse}

\textbf{Evaluation Goal.}
SkillsUse is an in-house benchmark for evaluating dynamic skill utilization in open-ended, tool-enabled tasks.
It treats each skill as a structured instruction document that encodes task-specific knowledge, procedural constraints, and recommended workflows.
The benchmark evaluates whether a model can identify the relevant skill from the task context, understand its documentation, follow the prescribed workflow, execute appropriate tools, and produce a deliverable that reflects skill-specific requirements.

SkillsUse is designed to go beyond final task success.
In addition to assessing whether the model completes the task, it examines whether the success can be attributed to correct skill usage.
Specifically, the benchmark measures whether the model accesses the target skill, avoids distractor skills, extracts key constraints and procedures, applies them during execution, and obtains traceable gains over a generic solution.

\textbf{Data Composition.}
SkillsUse consists of two complementary settings: No-attachment and Attachment, each containing 200 tasks, for a total of 400 tasks.
Each task is constructed with one target skill and several distractor skills in the workspace.
The user request describes the task goal in natural language without explicitly naming the target skill, requiring the model to identify the applicable skill from task context rather than relying on explicit skill invocation.

\begin{itemize}
    \item \textbf{No-attachment.}
    Evaluation focuses on basic skill discovery and procedural adherence.
    The model must infer the relevant skill from the task requirements, read its documentation, and follow its key workflow without relying on additional user-provided materials.

    \item \textbf{Attachment.}
    Each task further includes task-aligned user materials, such as files, data, text snippets, or other attachments.
    Evaluation focuses on joint reasoning over user materials, skill documentation, tool outputs, and intermediate observations, with greater emphasis on integrating task-specific context with skill-specific constraints during execution.
\end{itemize}

Together, these two settings cover both controlled skill-use scenarios and realistic deep research workflows.
Compared with smaller skill-oriented evaluation tasks, SkillsUse provides broader coverage in task quantity, openness, and workflow complexity.

\textbf{Evaluation Metrics.}
SkillsUse evaluates each trajectory along three dimensions: Result, Execution, and Skill Usage.
The metric set contains 12 fine-grained criteria that jointly assess final deliverable quality, agentic execution quality, and attributable skill utilization.

\begin{itemize}
    \item \textbf{Result.}
    This dimension measures the quality and completeness of the final output.
    It covers main-task completion, subtask coverage, satisfaction of hard constraints, adherence to the required output structure, readability, usability, executability, and deliverable-level validity.
    It captures the extent to which the model produces a practically useful response or artifact that satisfies the user request.

    \item \textbf{Execution.}
    This dimension measures the quality of the model's observable action process.
    It evaluates whether the trajectory forms a coherent observation-action loop, including effective use of intermediate observations, avoidance of ineffective repetition, recovery from errors or incomplete information, reasonable ordering of actions, and adherence to the workflow specified by the skill.
    It captures the stability and reliability of agentic execution beyond the final output alone.

    \item \textbf{Skill Usage.}
    This dimension measures whether task completion can be attributed to correct use of the target skill.
    It evaluates target-skill discovery and reading, avoidance of distractor skills, extraction of skill-specific constraints and procedures, incorporation of these requirements into tool calls and final outputs, and traceable improvement over a generic solution.
    It distinguishes general task-solving ability from skill-grounded task execution.
\end{itemize}

\textbf{Evaluation Protocol.}
SkillsUse adopts a structured LLM-as-a-Judge protocol rather than relying solely on a single pass rate.
For each trajectory, the interaction history is first parsed into structured evidence, including the user request, available skills, accessed skill documents, executed tool calls, intermediate observations, error-recovery behavior, and final output.
An LLM-based judge then scores the trajectory according to predefined rubrics over the 12 fine-grained metrics described above.

To improve the reliability of open-ended evaluation, the judge is required to ground each decision in observable trajectory evidence rather than relying only on the apparent quality of the final answer.
This design is particularly important for skill-use evaluation, where a correct final output does not necessarily indicate that the model has selected, read, or followed the target skill.
By separately scoring Result, Execution, and Skill Usage, SkillsUse provides more fine-grained diagnostic information, allowing failures to be attributed to final-output quality, agentic execution, or skill-utilization errors.

This protocol complements existing skill evaluations that mainly focus on deterministic verification, final pass rate, or skill retrieval within a specific framework.
SkillsUse instead evaluates the full skill-use process in realistic open-ended tasks, including skill selection, documentation reading, workflow adherence, tool execution, and final delivery.

\section{Case Study Figures}
\label{app:case_study}

To complement the qualitative analysis in Section~\ref{sec:case_study}, additional case study figures are presented across the five capability dimensions discussed in the main text. They illustrate representative execution processes and outputs under diverse deep research scenarios.

\begin{figure*}[h]
    \centering
    \includegraphics[width=\textwidth,height=0.9\textheight,keepaspectratio,trim= 0 35 0 37,clip]{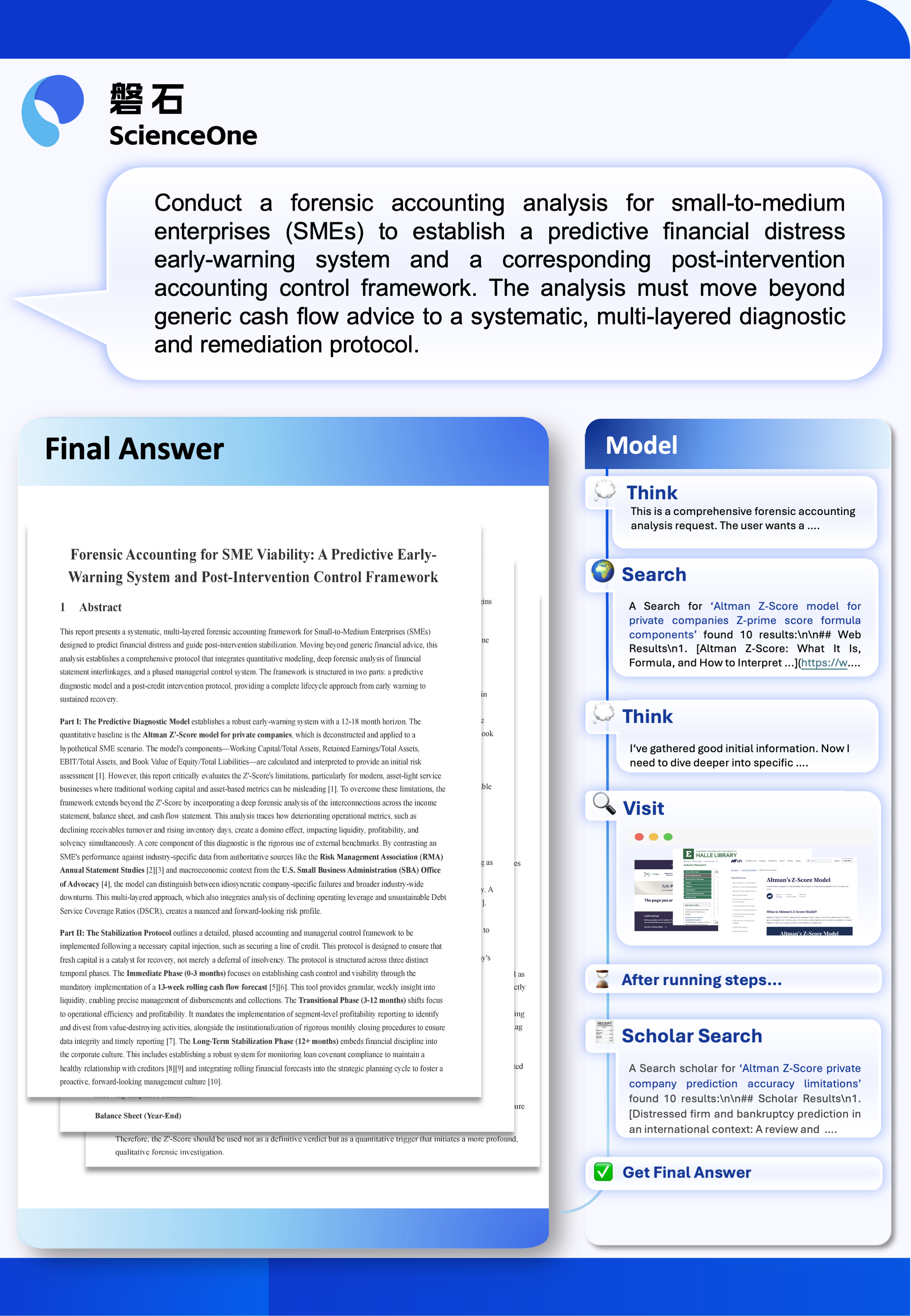}
    \caption{
    Deep research report generation case.
    Given an SME forensic accounting request, S1-DeepResearch constructs a structured financial distress early-warning system and a post-intervention accounting control framework, reflecting its ability to synthesize domain knowledge into a professional long-form report.
    }
    \label{fig:case_report_generation}
\end{figure*}

\begin{figure*}[p]
    \centering
    \includegraphics[width=\textwidth,height=0.9\textheight,keepaspectratio,trim= 0 35 0 37,clip]{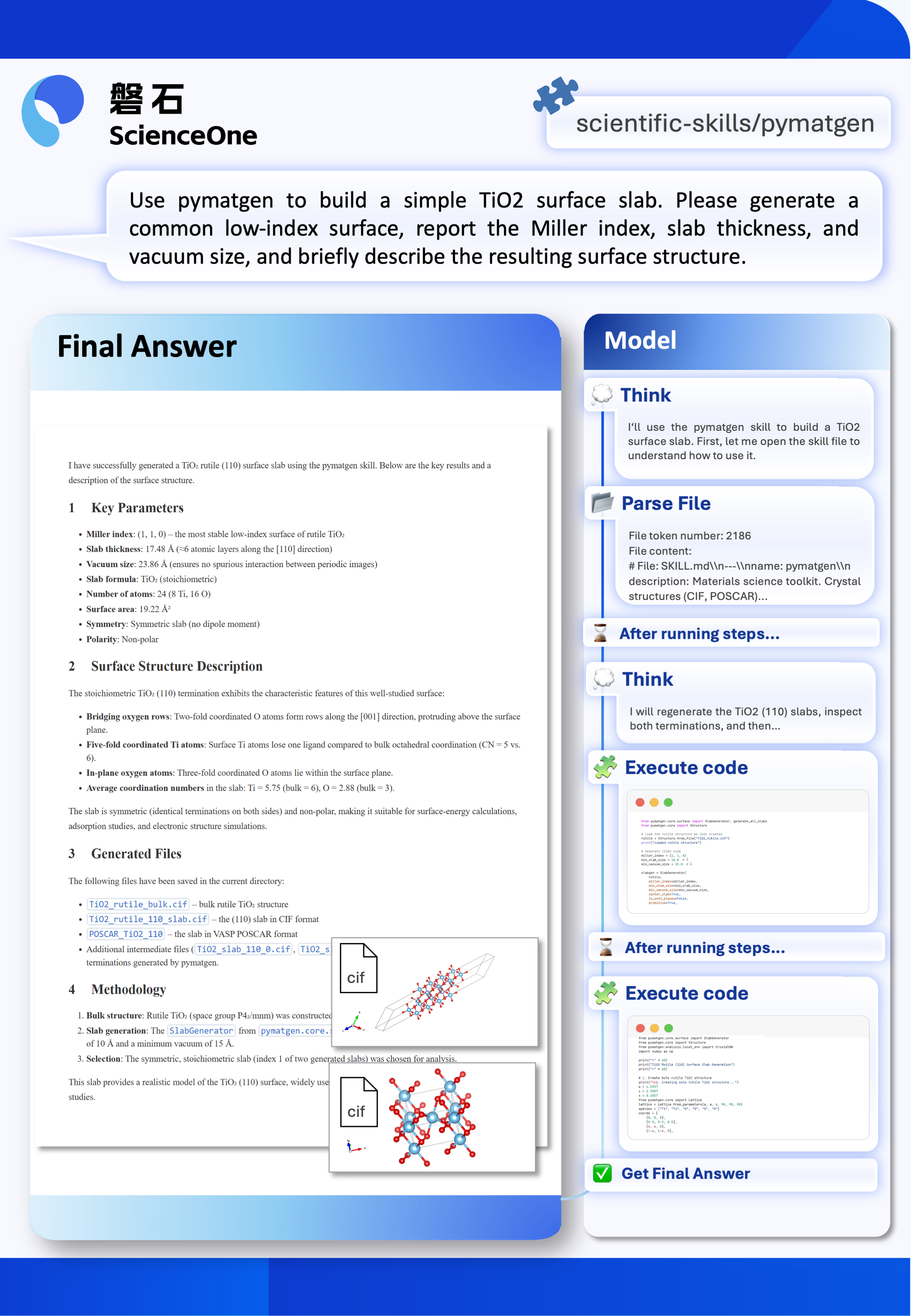}
    \caption{
    Dynamic skill utilization case.
    Given a TiO$_2$ surface modeling task, S1-DeepResearch uses scientific modeling tools to construct a rutile TiO$_2$ surface slab and report key structural properties, demonstrating its ability to coordinate domain knowledge with executable skills.
    }
    \label{fig:case_dynamic_skill}
\end{figure*}

\begin{figure*}[p]
    \centering
    \includegraphics[width=\textwidth,height=0.9\textheight,keepaspectratio,trim= 0 35 0 37,clip]{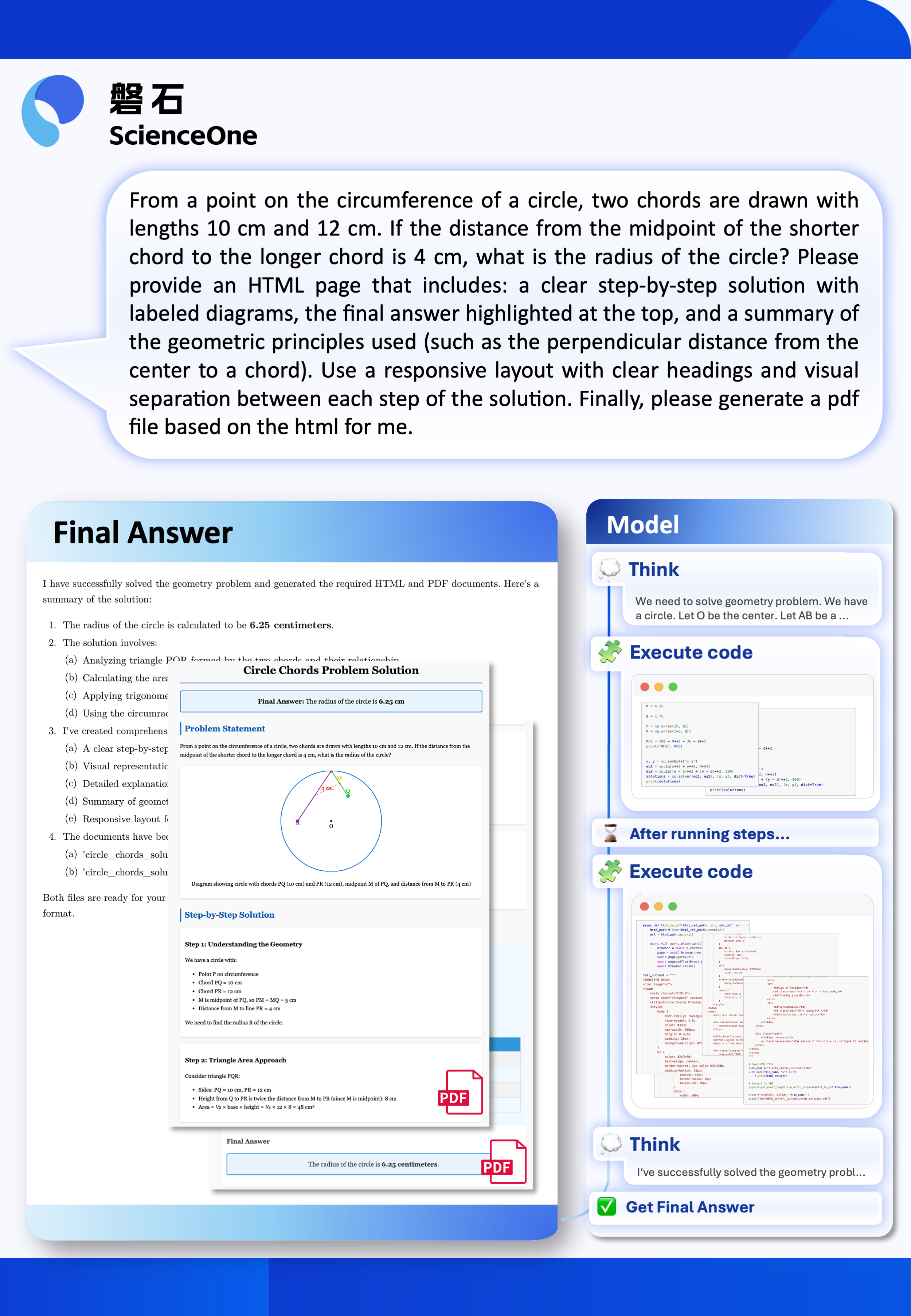}
    \caption{
    File understanding and generation case.
    Given a geometry problem with explicit document-generation requirements, S1-DeepResearch performs mathematical reasoning, organizes the solution into a structured HTML page, and generates a PDF artifact, showing its ability to connect reasoning with file-level output creation.
    }
    \label{fig:case_file_generation}
\end{figure*}

\begin{figure*}[p]
    \centering
    \includegraphics[width=\textwidth,height=0.9\textheight,keepaspectratio,trim= 0 35 0 37,clip]{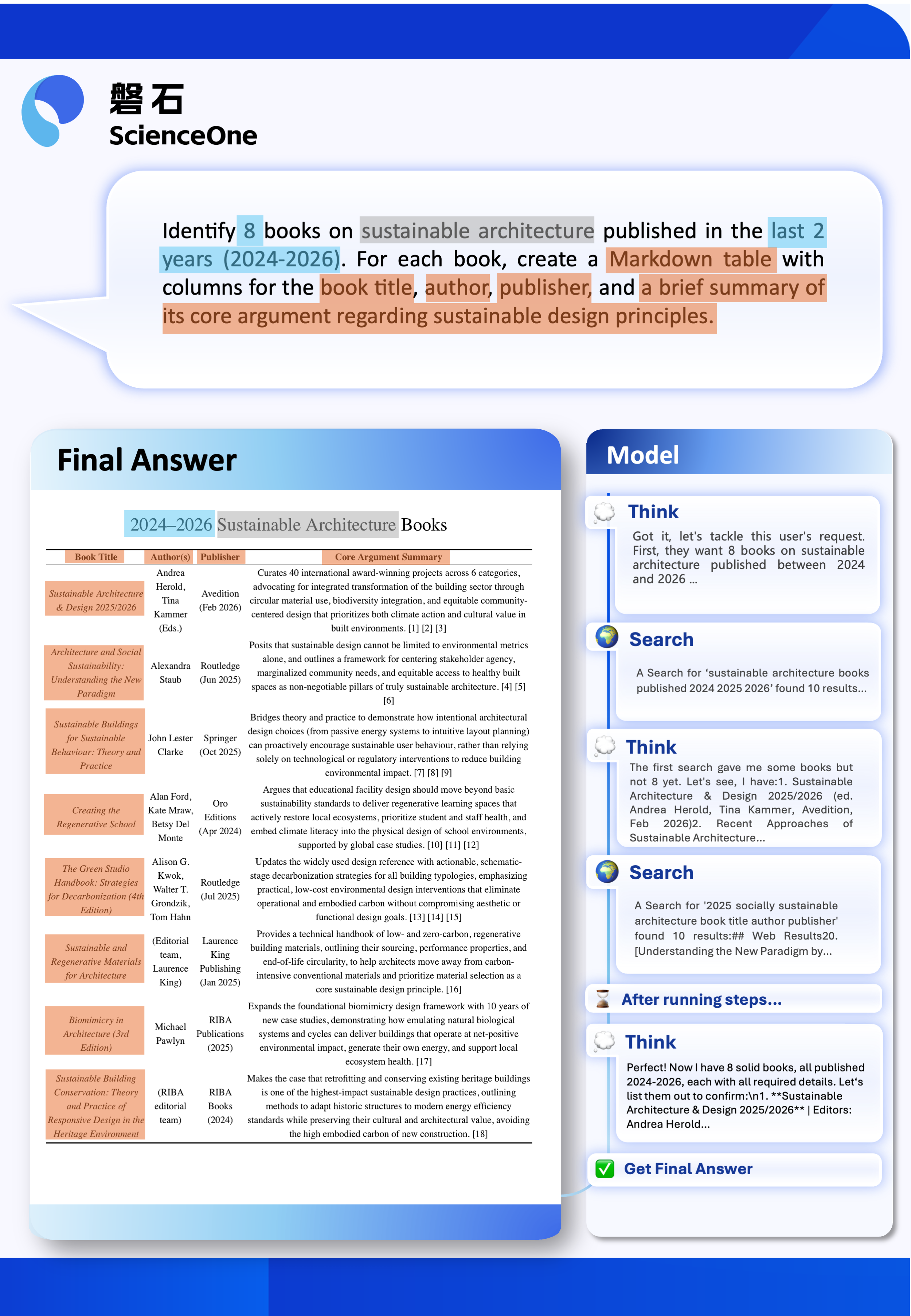}
    \caption{
    Deep research instruction following case.
    Given a constrained request for recent books on sustainable architecture, S1-DeepResearch identifies relevant publications, extracts required metadata, and formats the results as a Markdown table, demonstrating its ability to satisfy topical, temporal, and structural constraints.
    }
    \label{fig:case_instruction_books}
\end{figure*}

\begin{figure*}[p]
    \centering
    \includegraphics[width=\textwidth,height=0.9\textheight,keepaspectratio,trim= 0 35 0 37,clip]{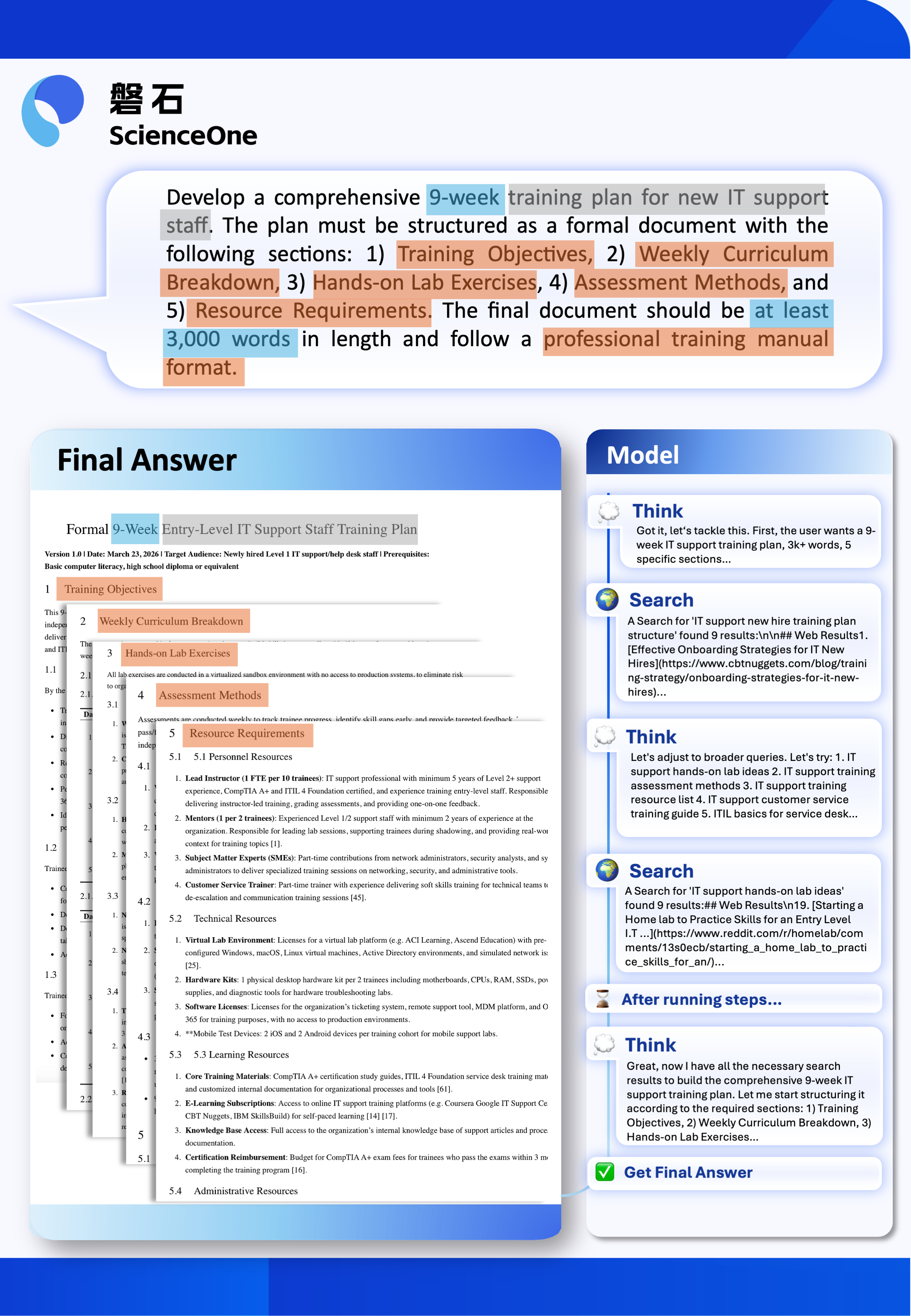}
    \caption{
    Long-form instruction following case.
    Given a request for a 9-week IT support training plan, S1-DeepResearch generates a professional training manual with specified sections, curriculum progression, hands-on exercises, assessment methods, and resource requirements.
    }
    \label{fig:case_instruction_training}
\end{figure*}

\begin{figure*}[p]
    \centering
    \includegraphics[width=\textwidth,height=0.9\textheight,keepaspectratio,trim= 0 35 0 37,clip]{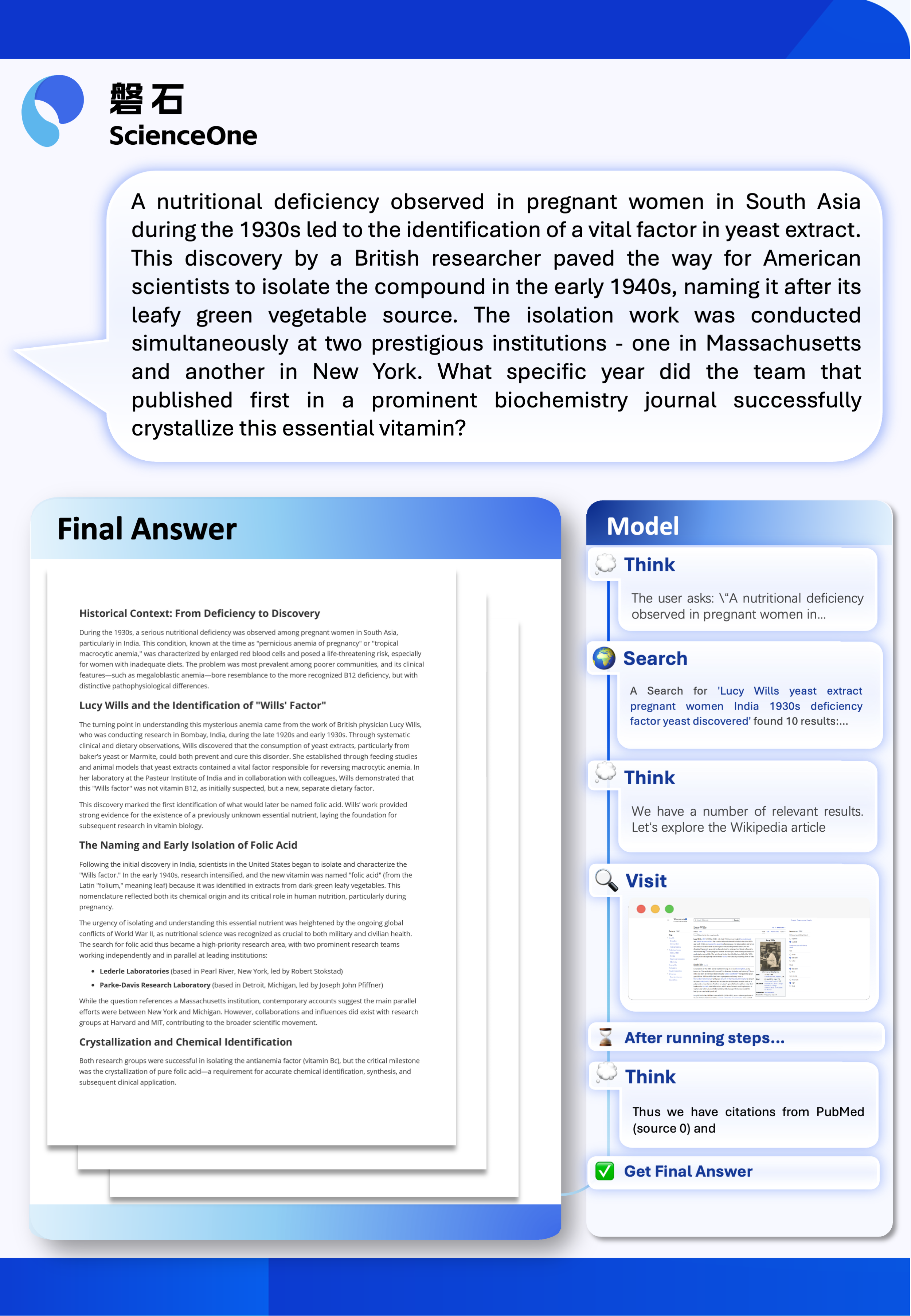}
    \caption{
    Long-horizon scientific reasoning case.
    Given a question about the discovery and crystallization of folic acid, S1-DeepResearch reconstructs the historical research timeline across medical observations, key researchers, institutional efforts, and publication records to identify the target year.
    }
    \label{fig:case_reasoning_folic}
\end{figure*}

\begin{figure*}[p]
    \centering
    \includegraphics[width=\textwidth,height=0.9\textheight,keepaspectratio,trim= 0 35 0 37,clip]{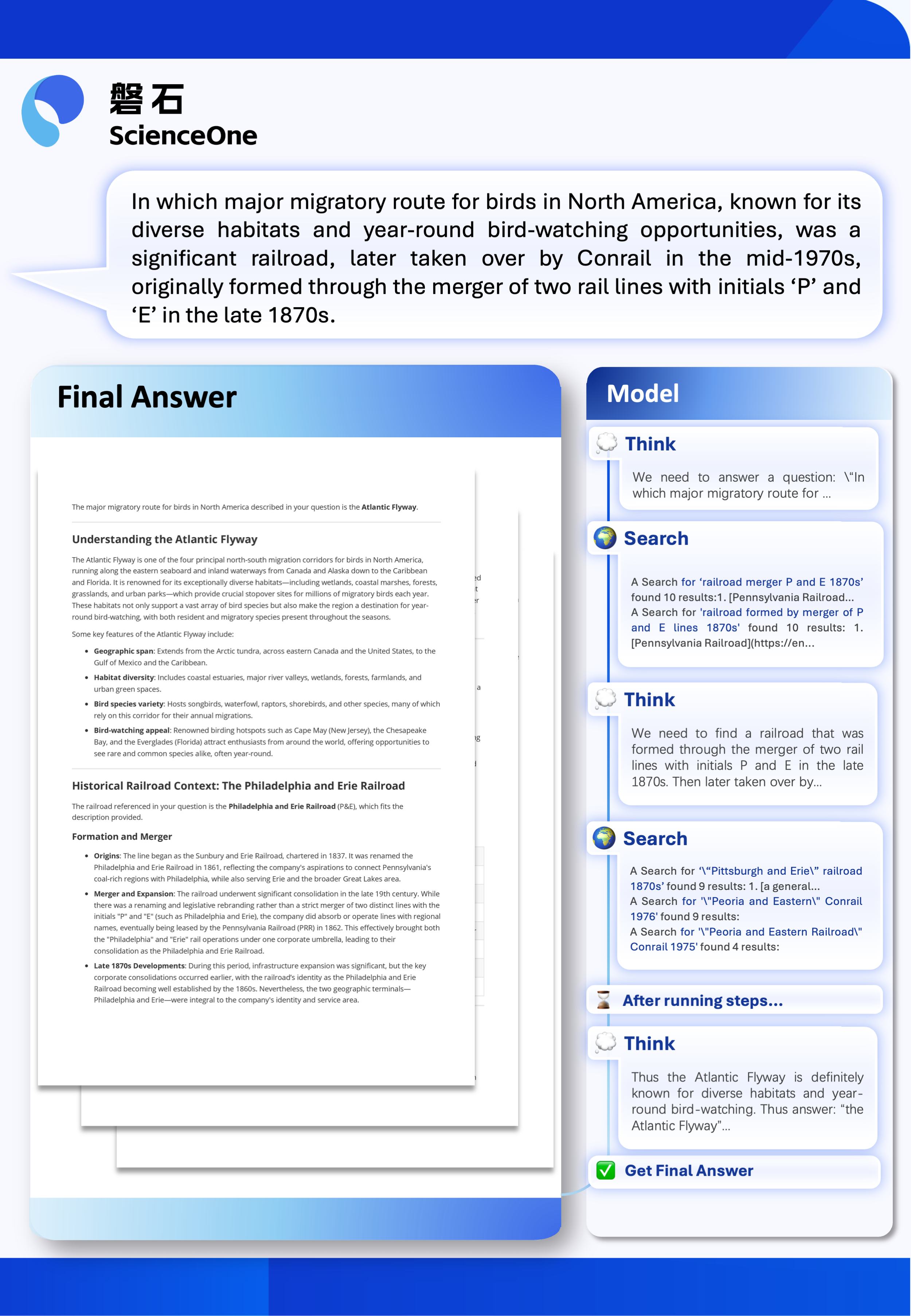}
    \caption{
    Cross-domain long-horizon reasoning case.
    Given a question linking bird migration routes with railroad history, S1-DeepResearch integrates ecological geography, regional transportation history, and corporate acquisition clues to infer the final answer.
    }
    \label{fig:case_reasoning_railroad}
\end{figure*}

\begin{figure*}[p]
    \centering
    \includegraphics[width=\textwidth,height=0.9\textheight,keepaspectratio,trim= 0 35 0 37,clip]{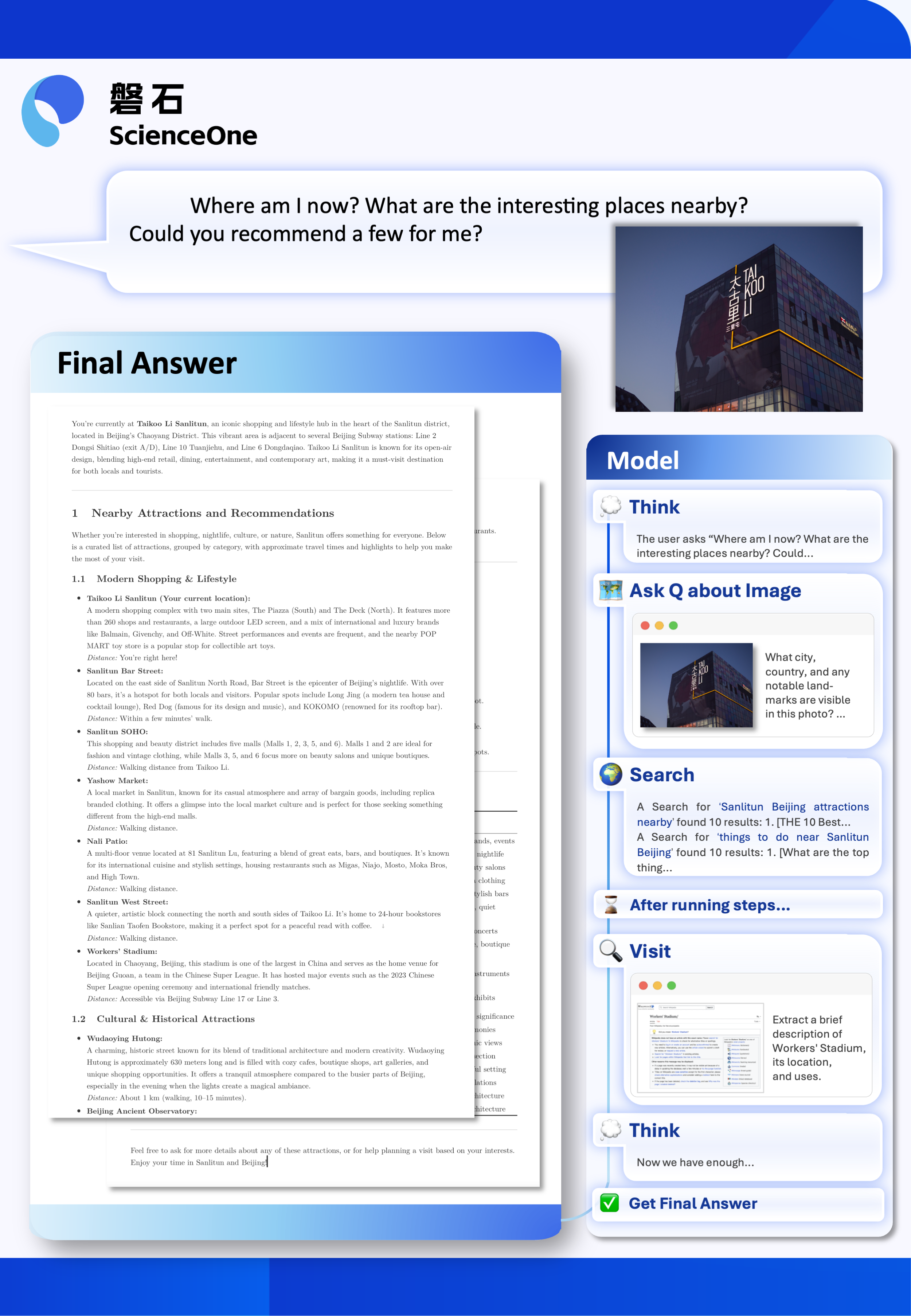}
    \caption{
    Multimodal long-horizon reasoning case.
    Given an image-based location and recommendation task, S1-DeepResearch infers the location from visual evidence, gathers external information about the surrounding area, and produces structured nearby-place recommendations.
    }
    \label{fig:case_reasoning_geo}
\end{figure*}

\end{document}